\documentclass[runningheads]{llncs}
\usepackage{graphicx}
\usepackage{comment}
\usepackage{amsmath,amssymb} 
\usepackage{multirow}
\usepackage{color}
\usepackage{adjustbox}
\usepackage[skip=2pt]{subcaption}
\usepackage{microtype} 
\usepackage{xspace}
\usepackage{makecell}
\usepackage{booktabs}
\usepackage{xspace}
\usepackage[dvipsnames]{xcolor}

\usepackage{wrapfig}

\def\xh{\hat x}
\def\yh{\hat y}
\def\zh{\hat y}

\newlength\secmargin
\newlength\paramargin
\newlength\figmargin

\begin{document}

\pagestyle{headings}
\mainmatter
\def\ECCVSubNumber{2233}  

\title{Predicting Camera Viewpoint Improves Cross-dataset Generalization for 
3D Human Pose Estimation}

\def\etal{\emph{et al.}\xspace}

\titlerunning{3D Human Pose Dataset Comparison}
%
\author{Zhe Wang \and Daeyun Shin \and Charless C. Fowlkes}
%
%

\institute{University of California, Irvine \\
\email{\tt\small \{zwang15,daeyuns, fowlkes\}@ics.uci.edu} 
\url{\tt\small \textcolor{red}{http://wangzheallen.github.io/cross-dataset-generalization}} 
}
\maketitle
\setlength{\figmargin}{-7.0mm}
\setlength{\paramargin}{-2.0mm}
\setlength{\secmargin}{-1.0mm}

\begin{abstract}
Monocular estimation of 3d human pose has attracted increased attention with
the availability of large ground-truth motion capture datasets.  However, the
diversity of training data available is limited and it is not clear to what
extent methods generalize outside the specific datasets they are trained on.
In this work we carry out a systematic study of the diversity and biases
present in specific datasets and its effect on cross-dataset generalization
across a compendium of 5 pose datasets. We specifically focus on systematic
differences in the distribution of camera viewpoints relative to a body-centered
coordinate frame.  Based on this observation, we propose an auxiliary task of
predicting the camera viewpoint in addition to pose. We find that models
trained to jointly predict viewpoint and pose systematically show significantly
improved cross-dataset generalization.
\end{abstract}

\section{Introduction}
\label{sec:introduction}

\begin{figure}[t]
\begin{center}
   \includegraphics[width=0.8\linewidth]{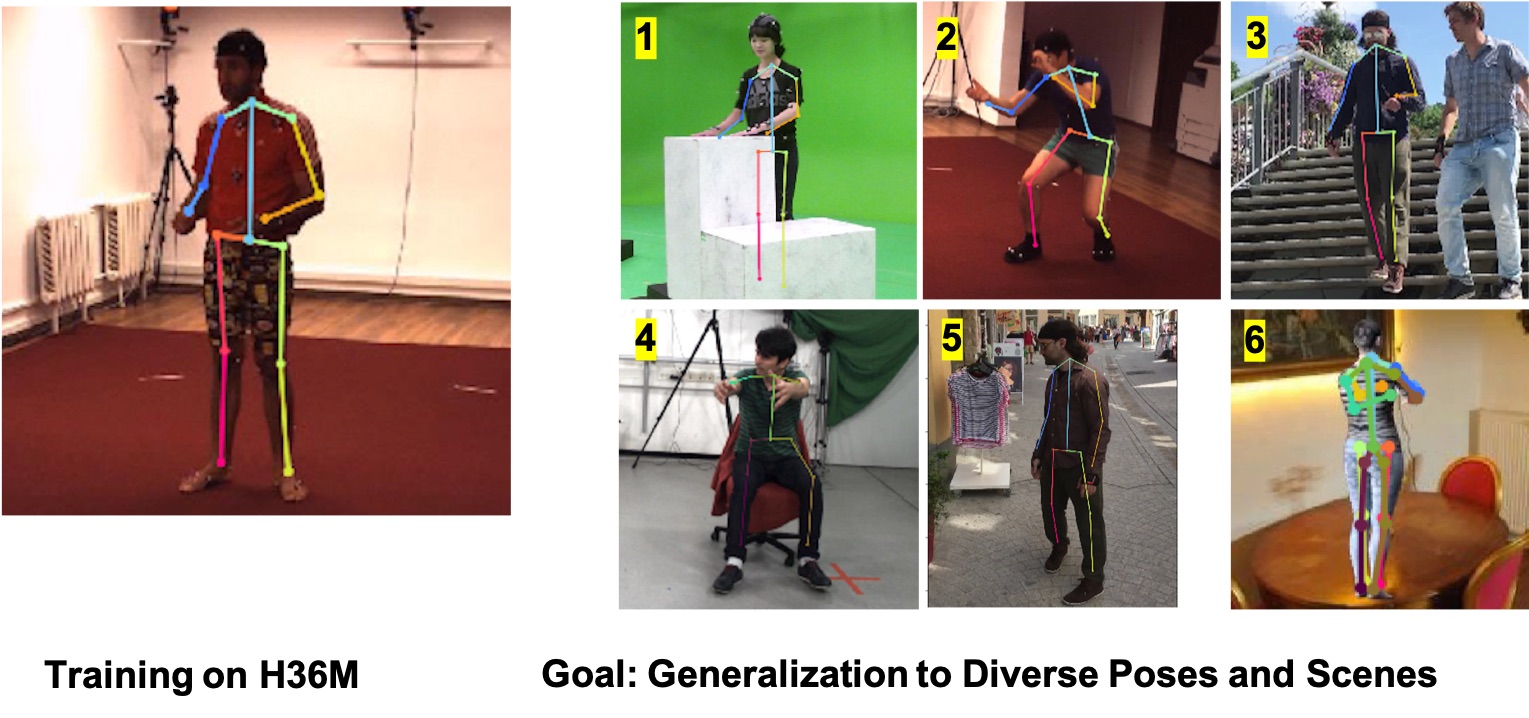}
\end{center}
   \caption[]{
   In this paper we consider the problem of dataset bias and cross-dataset generalization. Can you guess which human pose dataset each image on the
   right comes from? If we train a model on H36M data (left) can you predict which
   image has the lowest/highest 3D pose prediction error? (answer key below)\protect\footnotemark}
   
\label{fig:vis1sample}
\vspace{-0.15in}
\end{figure}

A large swath of computer vision research increasingly operates in playing
field which is swayed by the quantity and quality of annotated training data
available for a particular task. How well do you know your data?  Fig
\ref{fig:vis1sample} presents a sampling images from 5 popular datasets used
for training models for 3d human pose estimation (Human3.6M \cite{h36m_pami},
GPA \cite{gpa}, SURREAL \cite{varol17_surreal}, 3DPW \cite{inthewildeccv2018} ,
3DHP \cite{mono_3dhp2017}). We ask the reader to consider the game of ``Name
That Dataset'' in homage to Torralba~\etal{~\cite{unbiaseddataset}}.  Can you
guess which dataset each image belongs to? More importantly, if we train a
model on the Human3.6M dataset (at Fig \ref{fig:vis1sample} left) how well
would you expect it to perform on each of the images depicted?
 
Each of these datasets were collected using different mocap systems (VICON, The
Capture, IMU), different cameras (Kinect, commercial synchronized cameras,
phone), and collected in different environments (controlled lab environment,
marker-less in the wild environment, or synthetic images) with varying
camera viewpoint and pose distributions (see Fig \ref{fig:view_direction}).
These datasets contain further variations in body sizes, camera intrinsic and
extrinsic parameters, body and background appearance.  Despite the obvious
presence of such systematic differences, these variables and their subsequent
effect on performance have yet to be carefully analyzed. 

 \footnotetext{Answer key: Metric: MPJPE, the lower the better. 1) GPA: 69.7 mm
 2) H36M: 29.2 mm, 3) 3DPW, 71.2 mm, 4) 3DHP 107.7 mm, 5) 3DPW 66.2 mm, 6)
 SURREAL 83.4 mm, H36M image performs best while 3DHP image performs worst.}

In this paper, we study the generalization of 3d pose models across multiple
datasets and propose an auxiliary prediction task: estimating the relative rotation between camera viewing direction and a body-centered coordinate system defined by the orientation of the torso. This task serves to
significantly improve cross-dataset generalization. Ground-truth for our
proposed camera viewpoint task can be derived for existing 3D pose datasets
without requiring additional labels. We train off-the shelf models
\cite{rootnet,Zhou_2017_ICCV} which estimate the camera-relative 3d pose,
augmented with a viewpoint prediction branch. In our experiments, we show our
approach outperforms the state-of-the-art PoseNet \cite{rootnet} and
\cite{Zhou_2017_ICCV} baseline by a large margin across 5 different 3d pose
datasets. Perhaps even more startling is that the addition of this auxiliary
task results in significant improvement in cross-dataset test performance.
This simple approach increases robustness of the model and, to our knowledge,
is the first work that systematically confronts the problem of dataset 
bias in 3d human pose estimation.
\\

\noindent To summarize, our main contributions are:

$\bullet$ We analyze the differences among contemporary 3d human pose
estimation datasets and characterize the distribution and diversity of
viewpoint and body-centered pose.

$\bullet$ We propose the novel use of camera viewpoint prediction as an
auxiliary task that systematically improves model generalization by limiting
overfitting to common viewpoints and can be directly calculated from commonly
available joint coordinate ground-truth. 

$\bullet$ We experimentally demonstrate the effectiveness of the viewpoint
prediction branch in improving cross-dataset 3d human pose estimation over two
popular baseline and achieve state-of-the-art performance on five datasets. 

\section{Related Work}
\label{sec:relatedwork}

\paragraph{Cross-Dataset Generalization and Evaluation}
3d human pose estimation from monocular imagery has attracted significant
attention due to its potential utility in applications such as motion
retargeting \cite{motionretarget}, gaming, sports analysis, and health care
\cite{austim}.  Recent methods are typically based on deep neural network
architectures \cite{simple,rootnet,volumetric,pavllo:quaternet:2018,integral,Zhou_2017_ICCV}
trained on one of a few large scale, publicly available datasets. Among these
are \cite{simple,volumetric,integral} evaluated on H36M,
\cite{mono_3dhp2017,Zhou_2017_ICCV} work on both H36M \cite{h36m_pami} and 3DHP
\cite{mono_3dhp2017}, \cite{inthewildeccv2018,Trumble:BMVC:2017} work on
TOTALCAPTURE \cite{Trumble:BMVC:2017} and 3DPW\cite{inthewildeccv2018},
\cite{gpa} work on the GPA dataset \cite{gpa}. \cite{varol17_surreal} works on
both SURREAL \cite{varol17_surreal} and H36M \cite{h36m_pami} dataset. 

Given the powerful capabilities of CNNs to overfit to specific data, we are
inspired to revisit the work of \cite{unbiaseddataset}, which presented a
comparative study of popular object recognition datasets with the goals of
improving dataset collection and evaluation protocols. Recently, 
\cite{generalize_depth} observed characteristic biases present in commonly used
depth estimation datasets and proposed scale invariant training objectives to
enable mixing multiple, otherwise incompatible datasets.
\cite{generalize_handpose} introduced the first large-scale, multi-view
unbiased  hand pose dataset as training set to improve performance when testing
on other dataset.  Instead of proposing yet another dataset or resorting to
domain adaptation approaches (see e.g., \cite{universaldetection}), we focus
on identifying systematic biases in existing data and identifying generic
methods to prevent overfitting in 3d pose estimation.



\paragraph{Coordinate Frames for 3D Human Pose}
In typical datasets, gold-standard 3d pose is collected with motion capture
systems~\cite{h36m_pami,humaneva,Trumble:BMVC:2017,gpa} and used to define
ground-truth 3D pose relative one or more calibrated RGB camera coordinate
systems~\cite{h36m_pami,inthewildeccv2018,mono_3dhp2017,varol17_surreal,gpa}.
To generate regression targets for use in training and evaluation, it is
typical to predict the {\em relative} 3d pose and express the joint positions
relative to a specified root joint such as the pelvis (see
e.g.,\cite{rootnet,integral}).  We argue that camera viewpoint is an important
component of the experimental design which is often overlooked and explore 
using a body-centered coordinate system which is rotated relative to the 
camera frame.

This notion of view-point invariant prediction has been explored in the context
of 3D object shape estimation
\cite{choy20163d,groueix2018,mescheder2019occupancy,richter2018matryoshka,shin2018pixels,tatarchenko2017octree,drcTulsiani17,yan2016perspective}
where many works have predicted shape in either an object-centered  or camera-centered
coordinate frame \cite{shin2018pixels,tatarchenko2019single,zhang2018learning}. 
Closer to our task is the 3d hand pose
estimator of \cite{zb2017hand} which separately estimated the viewpoint and
pose (in canonical hand-centered coordinates similar to ours) and then combine
the two to yield the final pose in the camera coordinate frame. However, we
note that predicting canonical pose directly from image features is difficult
for highly articulated objects (indeed subsequent work on hand pose, e.g.
\cite{eccv2018handpose}, abandoned the canonical frame approach).  Our use of
body-centered coordinate frames differs in that we only use them as a
auxiliary training task that improves prediction of camera-centered pose. 


\paragraph{3D Human Pose Estimation} 
With the recent development of deep neural networks (CNNs),
there are  significant  improvements  on  3D  human  pose  estimation \cite{inthewildintermediate,simple,volumetric,xiao2018simple}. Many of them try to tackle in-the-wild images. \cite{Zhou_2017_ICCV} proposes to add bone length constraint to generalize their methods to in the wild image.  \cite{LCRnet++} seeks to pose anchors as classification template and refine the prediction with further regression loss. \cite{inthewildintermediate}  propose a a new disentangled hidden space
encoding of explicit 2D and 3D features for monocular 3D human pose estimation
that shows high accuracy and generalizes well to in-the-wild scenes, however,
they do not evaluate its capacity on indoor cross-dataset generalization. To
the best of our knowledge, our work is the first to exploit cross-dataset task
not only towards in-the-wild generalization but also across different indoor
datasets.


\paragraph{Multi-task Training}
There have has been a wide variety of work in training deep CNNs to perform multiple tasks, for example: joint detection, classification, and segmentation \cite{maskrcnn}, joint surface normal, depth, and semantic segmentation \cite{ubernet}, joint face detection, keypoint, head orientation and attributes \cite{hyperface}. 
Such work typically focuses on the benefits (accuracy and computation)
of jointly training a single model for two or more related tasks.
For example, predicting face viewpoint has been shown to improve face recognition \cite{yin2017multi}. Our approach
to improving generalization differs in that we train models to perform
two tasks (viewpoint and body pose) but discard viewpoint predictions at test 
time and only utilize pose. In this sense our model is more closely 
related to work on ``deeply-supervised'' nets \cite{lee2015deeply,HEG} which 
trains using losses associated with auxiliary branches that are not  
used at test time.

\begin{table}[t] \centering
\begin{center}
{\scriptsize
\begin{tabular}{@{}l c c c c c@{}}
\toprule
Dataset & H36M & GPA & SURREAL & 3DPW & 3DHP \\
\hline
Year & 2014 & 2019 & 2017 & 2018 & 2017 \\
Imaging Space & 1000 $\times$ 1002 & 1920 $\times$ 1080 & 320 $\times$ 240 & 1920 $\times$ 1080 & 2048 $\times$ 2048 \\ 
 &   &   &   &  & or 1920 $\times$ 1080 \\ 
Camera Distance & 5.2 $\pm$ 0.8 & 5.1 $\pm$ 1.2 & 8.0 $\pm$ 1.0 & 3.5 $\pm$ 0.7 & 3.8 $\pm$ 0.8 \\
Camera Height & 1.6 $\pm$ 0.05 & 1.0 $\pm$ 0.3 & 0.9 $\pm$ 0.1 & 0.6 $\pm$ 0.8 & 0.8 $\pm$ 0.4 \\
Focal Length& 1146.8 $\pm$ 2.0 & 1172.4 $\pm$ 121.3 & 600 $\pm$ 0 & 1962.2 $\pm$ 1.5 & 1497.88 $\pm$ 2.8 \\
No. of Joints & 38 & 34 & 24 & 24 & 28 or 17 \\
No. of Cameras & 4 & 5 & 1 & 1 & 14 \\
No. of Subjects & 11 & 13 & 145 & 18 & 8 \\
Bone Length & 3.9 $\pm$ 0.1 & 3.7 $\pm$ 0.2 & 3.7 $\pm$ 0.2 & 3.7 $\pm$ 0.1 & 3.7 $\pm$ 0.1 \\
GT source &VICON & VICON & Rendering &  SMPL & The Capture \\
No. Train Images & 311,951 & 222,514 &   867,140 & 22,375 & 366,997 \\
No. Test Images  & 109,764 & 82,378 & 507 & 35,515 &  2,875\\
\bottomrule
\end{tabular}
}
\end{center}

\caption{Comparison of existing datasets commonly used for training and
evaluating 3D human pose estimation methods. We calculate the mean and std of
camera distance, camera height, focal length, bone length from training set.
Focal length is in mm while the others are in unit meters. 
3DHP has two kinds of cameras and the training set
provide 28 joints annotation while test set provide 17 joints annotation.}
\label{table:datasets}
\vspace{-0.15in}
\end{table}

\section{Variation in 3D Human Pose Datasets}
\label{sec:datasetanalysis}


We begin with a systematic study of the differences and biases across 3d pose
datasets.  We selected three well established datasets Human3.6m (H36M),
MPI-inf-3dhp (3DHP), SURREAL, as well as two more recent datasets 3DPW and GPA
for analysis.  These are large-scale datasets with a wide variety of
characteristics in terms of capture technology, appearance
(in-the-wild,in-the-lab,synthetic) and content (range of body sizes, poses,
viewpoints, clothing, occlusion and human-scene interaction).  In this paper,
we focus on characterizing variation in geometric quantities (pose and viewpoint)
which can be readily quantified (compared to, e.g., lighting and clothing).

We list some essential statistics from 5 datasets in Table \ref{table:datasets}. 
For these datasets, gold-standard 3d pose is collected with motion capture
systems~\cite{h36m_pami,humaneva,Trumble:BMVC:2017,gpa} and used to define
ground-truth 3D pose relative one or more calibrated RGB camera coordinate
systems~\cite{h36m_pami,inthewildeccv2018,mono_3dhp2017,varol17_surreal,gpa}.
To generate regression targets for use in training and evaluation, it is
typical to predict the {\em relative} 3d pose (see
e.g.,\cite{rootnet,integral}) and express the joint positions relative to a
specified root joint (typically the pelvis) and crop/scale the input image
accordingly.  This pre-processing serves to largely ``normalize away'' dataset
differences in camera intrinsic parameters and camera distance shown in Table
\ref{table:datasets}.  However, it does not address camera orientation.

\begin{wrapfigure}[20]{r}{0.5\textwidth}
\vspace{-0.5in}
\begin{center}
\includegraphics[width=0.5\textwidth]{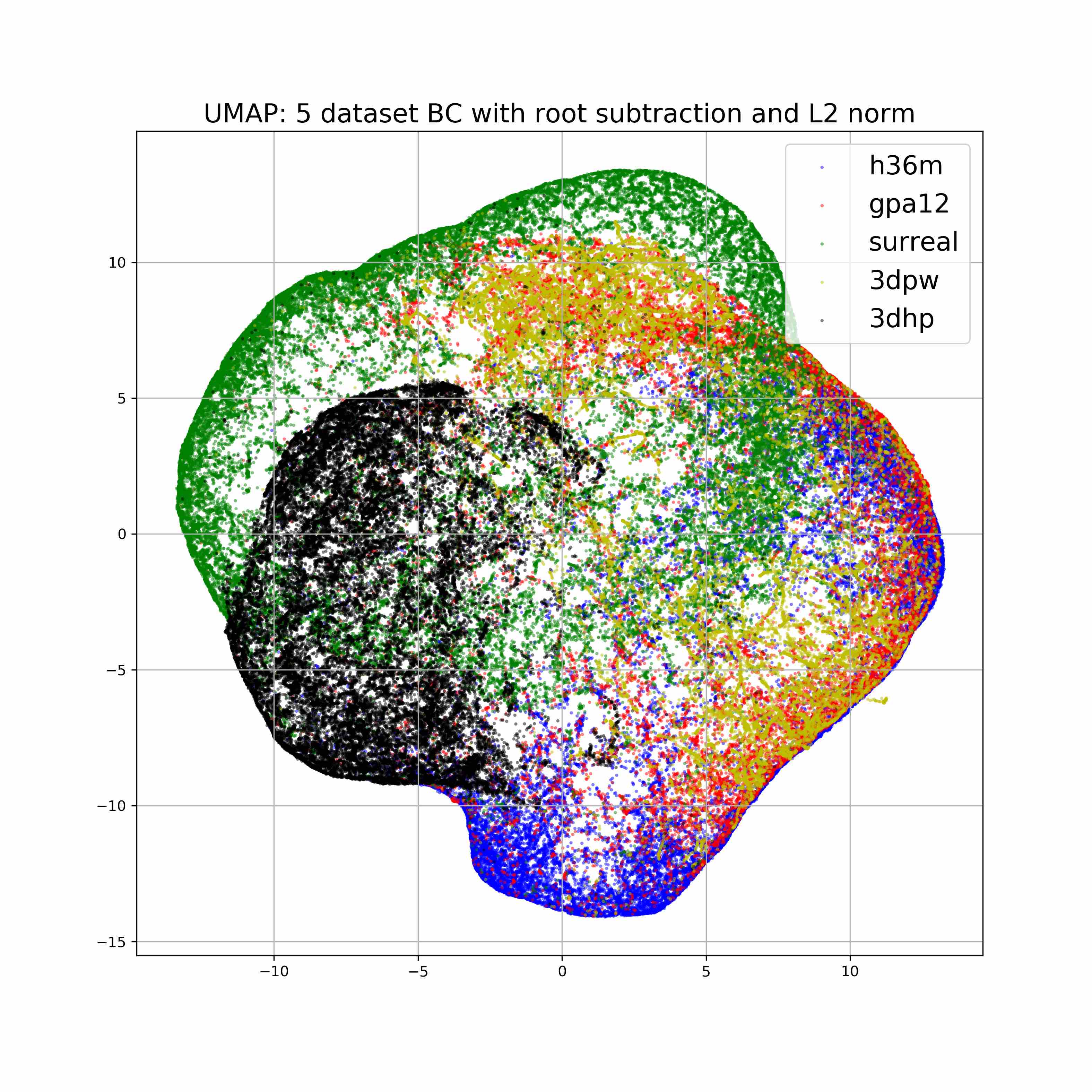}
\caption{\small Distribution of view-independent body-centered pose, visualized 
   as a 2D embedding produced with UMAP \cite{mcinnes2018umap-software}}
   \label{fig:bclaumap}
\end{center}
\end{wrapfigure}

To characterize the remaining variability, we factor the camera-relative
pose into camera viewpoint (the position of the camera relative to a canonical
body-centered coordinate frame defined by the orientation of the person's
torso) and the pose relative to this body-centered coordinate frame.


\paragraph{Computing Body-centered Coordinate Frames}
To define a
viewpoint-independent pose, we need to specify a canonical body-centered
coordinate frame. As shown in Fig \ref{fig:2a}, we take the origin to be the 
camera-centered coordinates of root joint (pelvis) $p_p = (x_p,y_p,z_p)$ and
the orientation is defined by the plane spanned by $p_p$, the left shoulder
$p_l$ and the right shoulder $p_r$. Given these joint positions, we can compute
an orthogonal frame consisting of the front direction $f$, up direction $u$ and
right direction $r$ are defined as:
\begin{align*} 
u &=  (p_l + p_r) / 2 - p_p\\
f &= (p_l - p_p) \times (p_r - p_p) \\
r &=  f  \times u  
\end{align*}
The rotation between the body-centered frame and the camera frame is then given
by the matrix $R = -[r,u,f]$. We find it useful to represent rotations using
unit quaternions (as have others, e.g.
\cite{motionretarget,factored3dTulsiani17}). The corresponding unit quaternion
representing $R$ has components:
\begin{align} 
\label{eqn:matrixtoquaternion}
q &= \frac{1}{4q_0} [4 q_0^2, u_{2}-f_{1}, f_{0}-r_{2}, r_{1}-u_{0}],
&
q_0 = \sqrt{(1  - r_{0}  - u_{1} - f_{2} )}
\end{align}


\begin{figure}[t]
\begin{center}
\begin{subfigure}{0.48\textwidth}
\includegraphics[width=\linewidth]{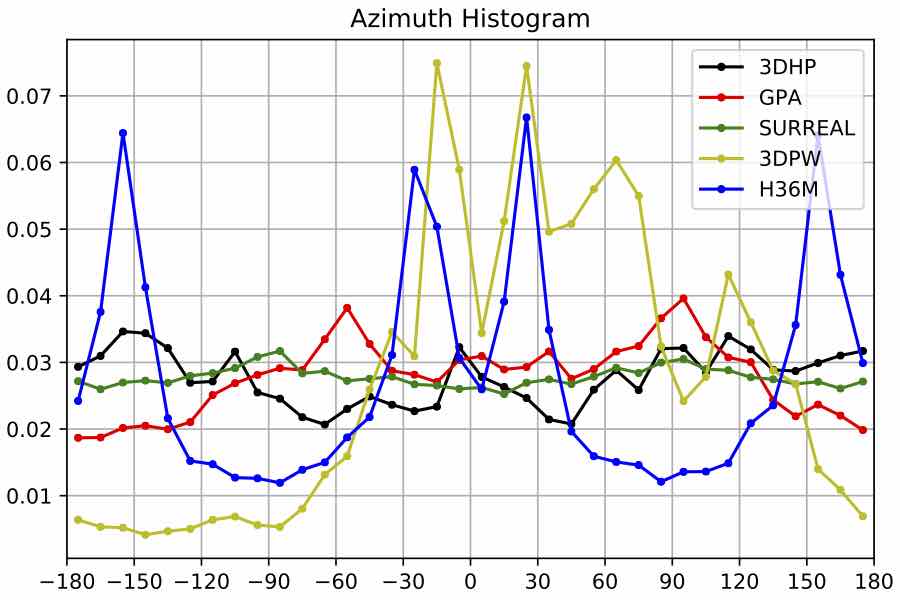}
\caption{Viewpoint Azimuth} \label{fig:az}
\end{subfigure}
\hspace*{\fill}
\begin{subfigure}{0.48\textwidth} \centering
\includegraphics[width=\linewidth]{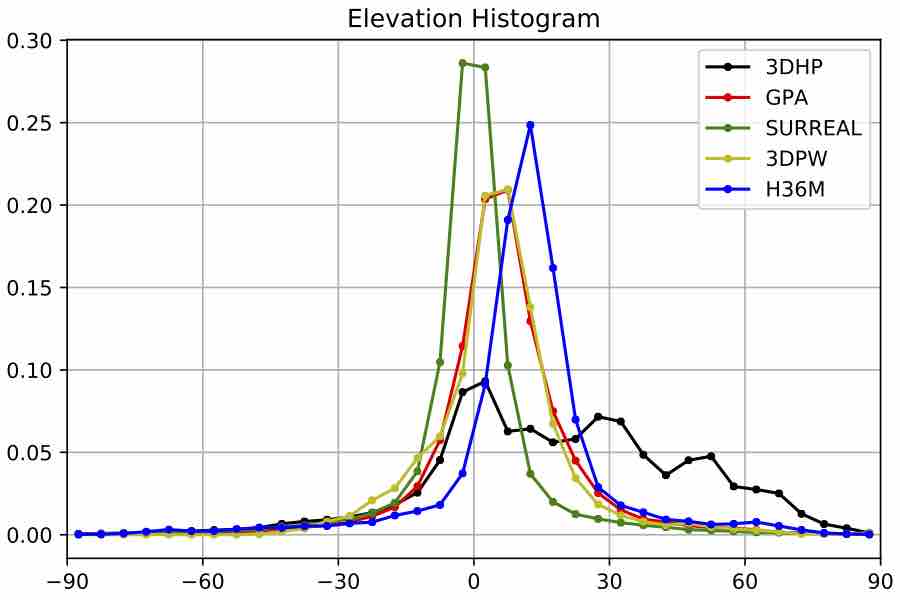}
\caption{Viewpoint Elevation} \label{fig:el}
\end{subfigure}
\end{center}
   \caption{
   Distribution of camera viewpoints relative to the human subject. 
   We show the distribution of camera azimuth $(-180^\circ,180^\circ)$ and
   elevation $(-90^\circ,90^\circ)$ for 50k poses sampled from each
   representative dataset
   (\textcolor{blue}{\textbf{H36M}}, \textcolor{red}{\textbf{GPA}},  \textcolor{OliveGreen}{\textbf{SURREAL}},
   \textcolor{Goldenrod}{\textbf{3DPW}}, \textcolor{black}{\textbf{3DHP}}).
   } 
\label{fig:view_direction}
\vspace{-0.15in}
\end{figure}

\paragraph{Distribution of Camera Viewpoints}
Fig \ref{fig:view_direction} shows histograms capturing the distribution of
camera viewing direction in terms of azimuth (Fig \ref{fig:az}) and elevation
(Fig \ref{fig:el}) relative to the body-centered coordinate system for 50k
sample poses from each of the 5 datasets.

We observe \textbf{H36M} has a wide range of view direction over azimuth with
four distinct peaks ($-30$ degree, 30 degree, $-160$ degree, 160 degree), it
shows during the capture session subjects are always facing towards or facing
away the control center while the four RGB cameras captured from four corners.
H36M has a clear bias towards elevation above 0; \textbf{GPA} is more spread
over azimuth compared with H36M, most of the views range from $-60$ degree to
90 degree; 
\textbf{SURREAL} synthetically sampled camera positions with a uniform
distribution over azimuth, and also have a uniform 
distribution over elevation. The viewpoint bias for \textbf{3DPW} arises 
naturally from filming people in-the-wild from a handheld or tripod mounted
camera roughly the same height as the subject.
Of the non-synthetic datasets, \textbf{3DHP} is the most uniform spread over
azimuth and includes a wider range of positive elevations, a result of
utilizing cameras mounted at multiple heights including the ceiling. 

These differences are further highlighted in Fig \ref{fig:body-centeredUFR}
which shows the joint distribution of camera views and reveals the source of
non-uniformity of the azmuthal distribution for 3DHP and H36M due to subjects
tending to face a canonical direction while performing some actions.  For
example, in H36M in Fig \ref{fig:2b}, actions in which the subject lean over or
lie down (extreme elevations) only happen at particular azimuths.  Similarly,
in 3DHP (Fig \ref{fig:2f}), the 14 camera locations are visible as dense
clusters at specific azimuths indicating a significant subset of the data in
which the subject was facing in a canonical direction relative to the camera
constellation.

\paragraph{Distribution of Pose}
To characterize the remaining variability in pose after the viewpoint is 
factored out, we used the coordinates of 14 joints common to all datasets 
expressed in the body-centered coordinate frame. We also scaled the
body-centered joint locations to a common skeleton size (removing variation in
bone length shown in Table 1). To visualize the resulting high-dimensional 
data distribution, we utilized UMAP \cite{mcinnes2018umap-software} to perform a non-linear 
embedding into 2D.  Figure \ref{fig:bclaumap} shows the resulting distributions
which show a substantial degree of overlap. For comparison, please see 
the Appendix which show embeddings of the same data when bone length
and/or viewpoint are not factored out.  

We also trained a multi-layer perceptron to predict which dataset a
given body-relative pose came from. It had an average test accuracy of 20\%
providing further evidence of relatively little bias in the distribution of
poses across datasets once viewpoint and body size are factored out.

\section{Learning Pose and Viewpoint Prediction}
\label{sec:bcframe}

To overcome biases in viewpoint across datasets, we propose to use viewpoint
prediction as an auxiliary task to regularize the training of standard
camera-centered pose estimation models.

\subsection{Baseline architecture}
Our baseline model \cite{rootnet,Zhou_2017_ICCV}  consists of two parts: the
first ResNet \cite{resnet} backbone which takes in images patches cropped
around the human; followed by the second part which takes the resulting feature
map and upsamples it using three consecutive deconvolutional layers with batch
normalization and ReLU.  A 1-by-1 convolution is applied to the upsampled
feature map to produce the 3D heatmaps for each joint location.  The
soft-argmax \cite{integral} operation is used to extract the 2D image
coordinates $(\xh_j , \yh_j)$ of each joint $j$ within the crop, and the
root-relative depth $\zh_j$.  At test time, we can convert this prediction into
into a 3d metric joint location $p_j = (x_j,y_j,z_j)$ using the crop bounding
box, an estimate of the root joint depth or skeleton size, and the camera
intrinsic parameters.

\begin{figure}[t]
\begin{center}
   \includegraphics[width=\linewidth]{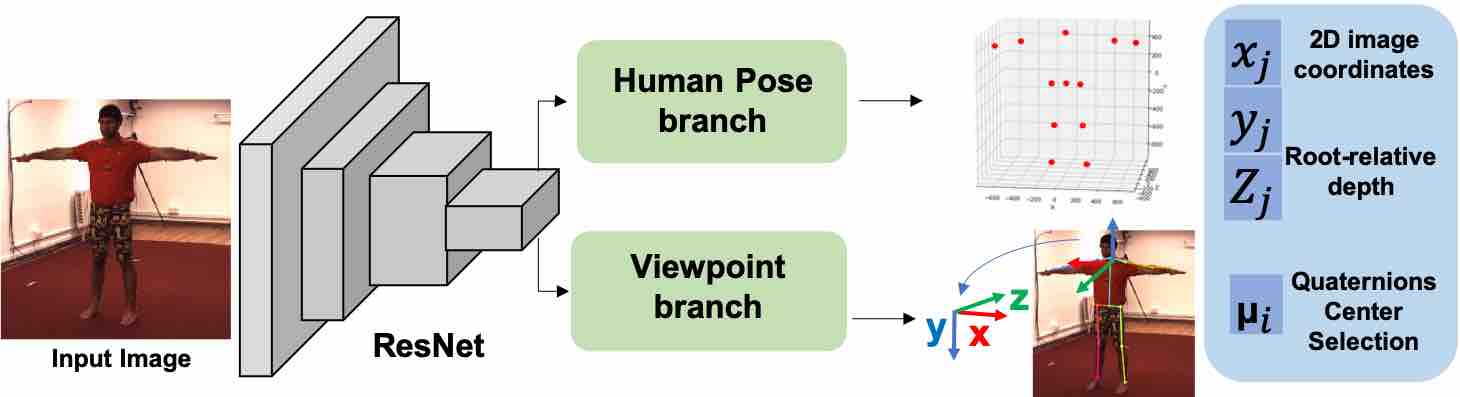}
\end{center}
   \caption{Flowchart of our model. 
   We augment a model which predicts camera-centered 3d pose using the
   \textbf{human pose branch} with an additional \textbf{viewpoint branch} 
   that selections among a set of quantized camera view directions.
   }
\label{fig:posenetflowchart}
\vspace{-0.15in}
\end{figure}

The loss function of the coordinate branch is the $L1$ distance between the
estimated and groud-truth coordinates. 
\begin{equation*}
  {\mathcal{L}_{\textit{pose}} =   \frac{1}{J}\sum^J_{j=1} ||p_j - p_j^{*}||_1 }
  \label{eqn:pose}
\end{equation*}


\begin{figure*}[t]
\centering
\begin{subfigure}{0.36\textwidth}
\includegraphics[width=\linewidth]{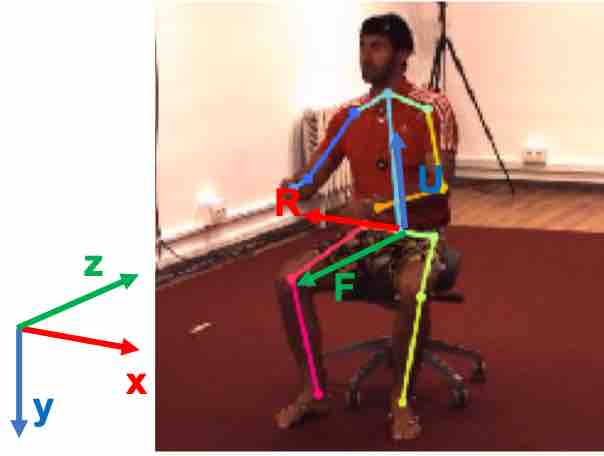}
\caption{Body-centered coordinate} \label{fig:2a}
\end{subfigure}
\hspace*{\fill}
\begin{subfigure}{0.30\textwidth}
\includegraphics[width=\linewidth]{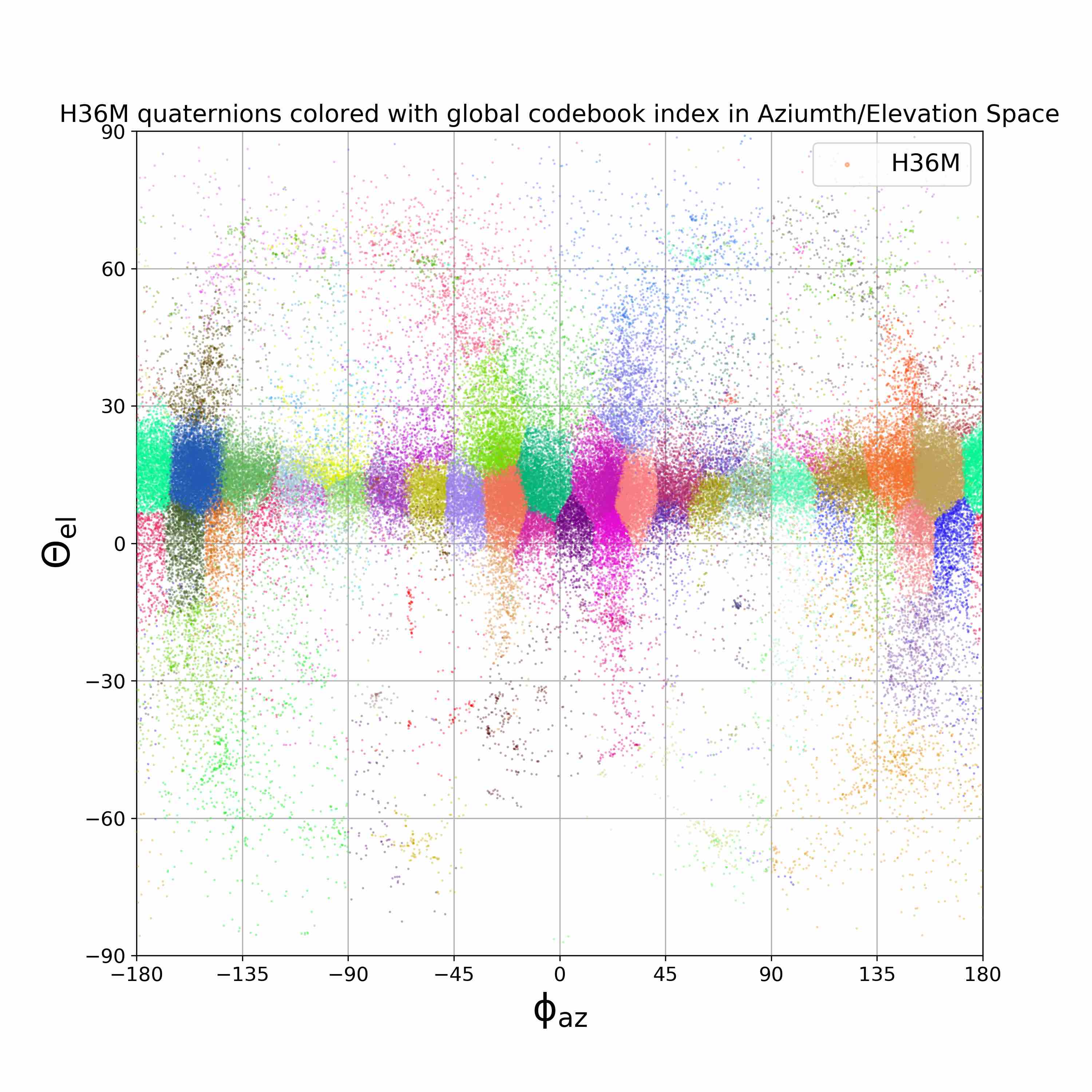}
\caption{\textsc{H36M}} \label{fig:2b}
\end{subfigure}
\hspace*{\fill}
\begin{subfigure}{0.30\textwidth}
\includegraphics[width=\linewidth]{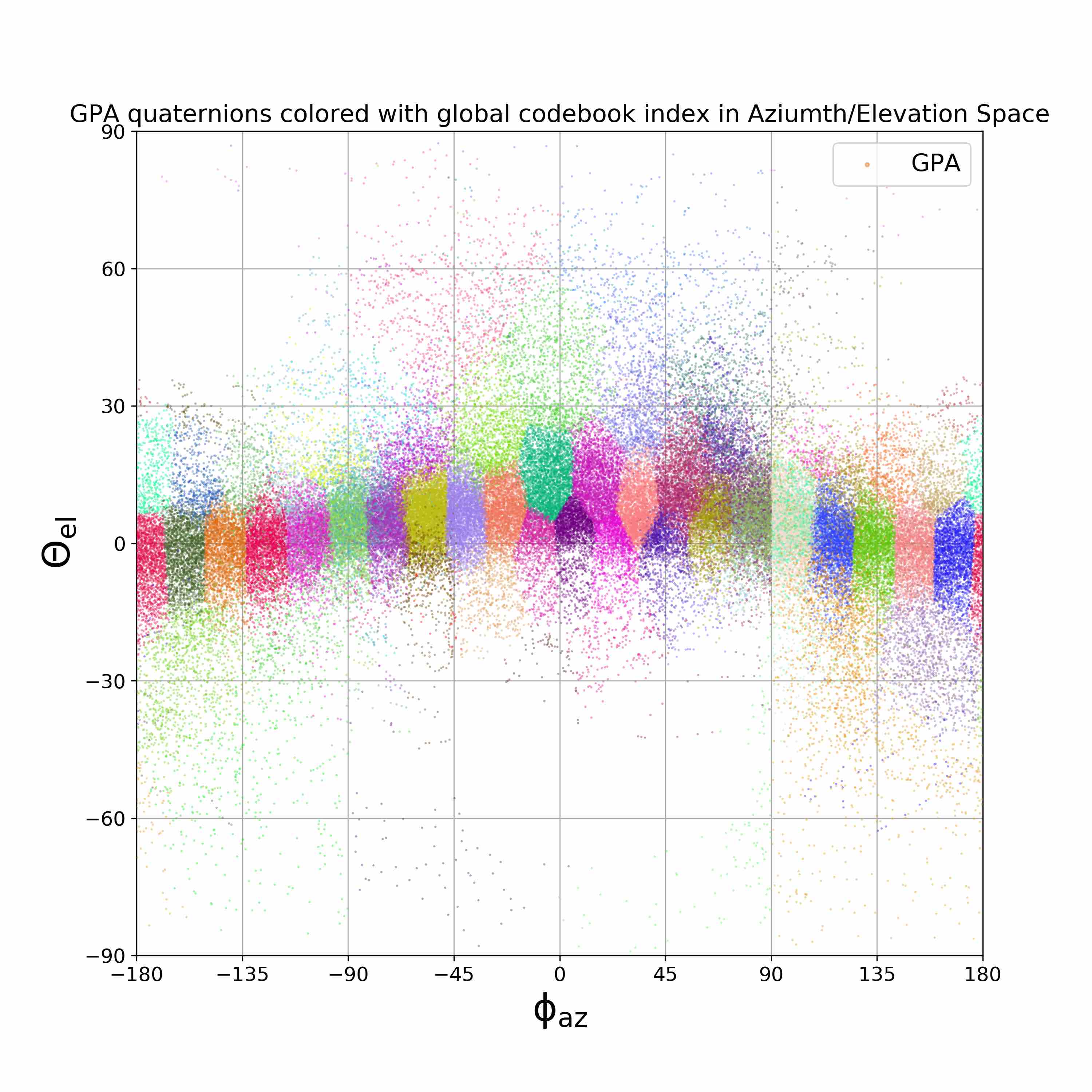}
\caption{\textsc{GPA}} \label{fig:2c}
\end{subfigure}
\hspace*{\fill}
\begin{subfigure}{0.32\textwidth}
\includegraphics[width=\linewidth]{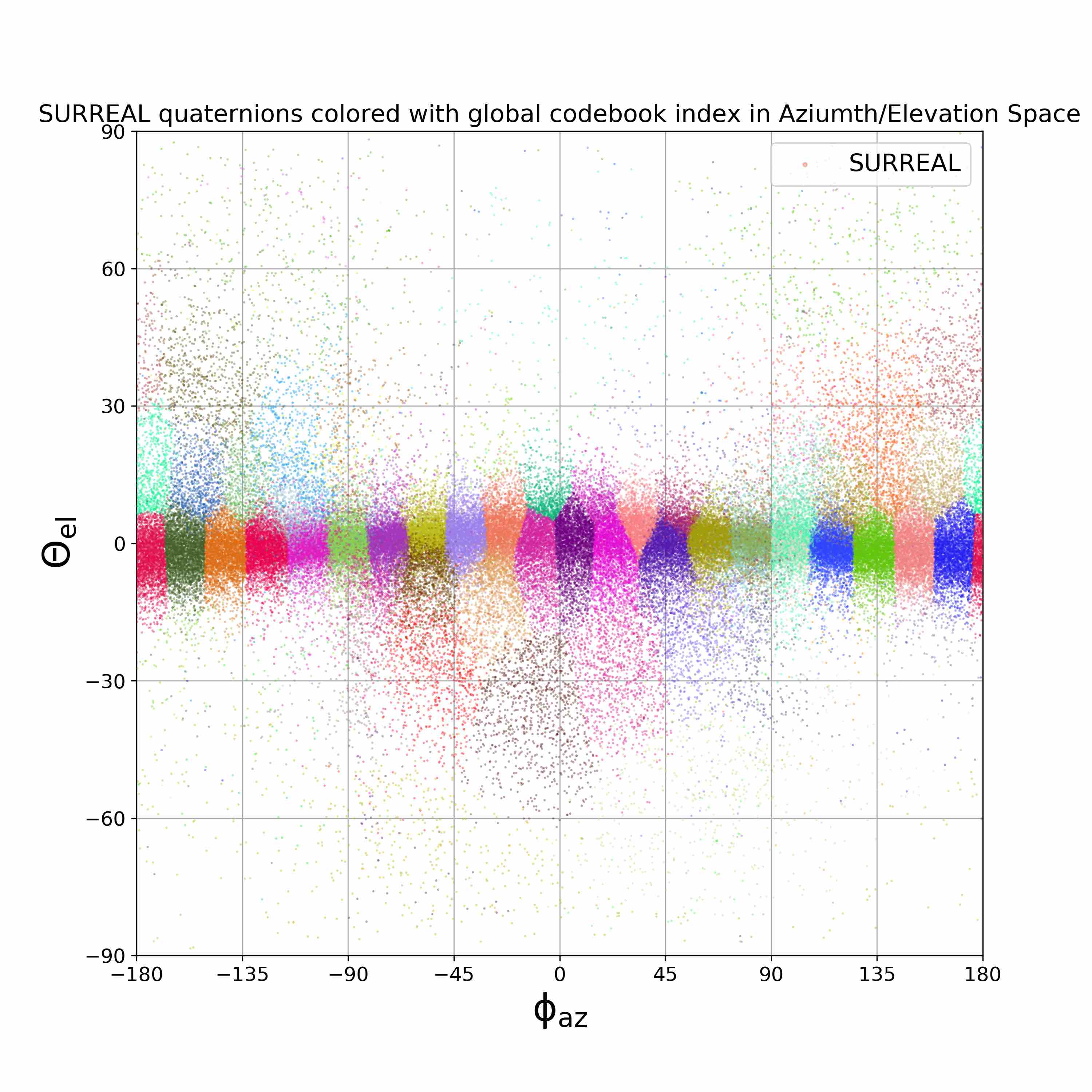}
\caption{\textsc{SURREAL}} \label{fig:2d}
\end{subfigure}
\hspace*{\fill}
\begin{subfigure}{0.32\textwidth}
\includegraphics[width=\linewidth]{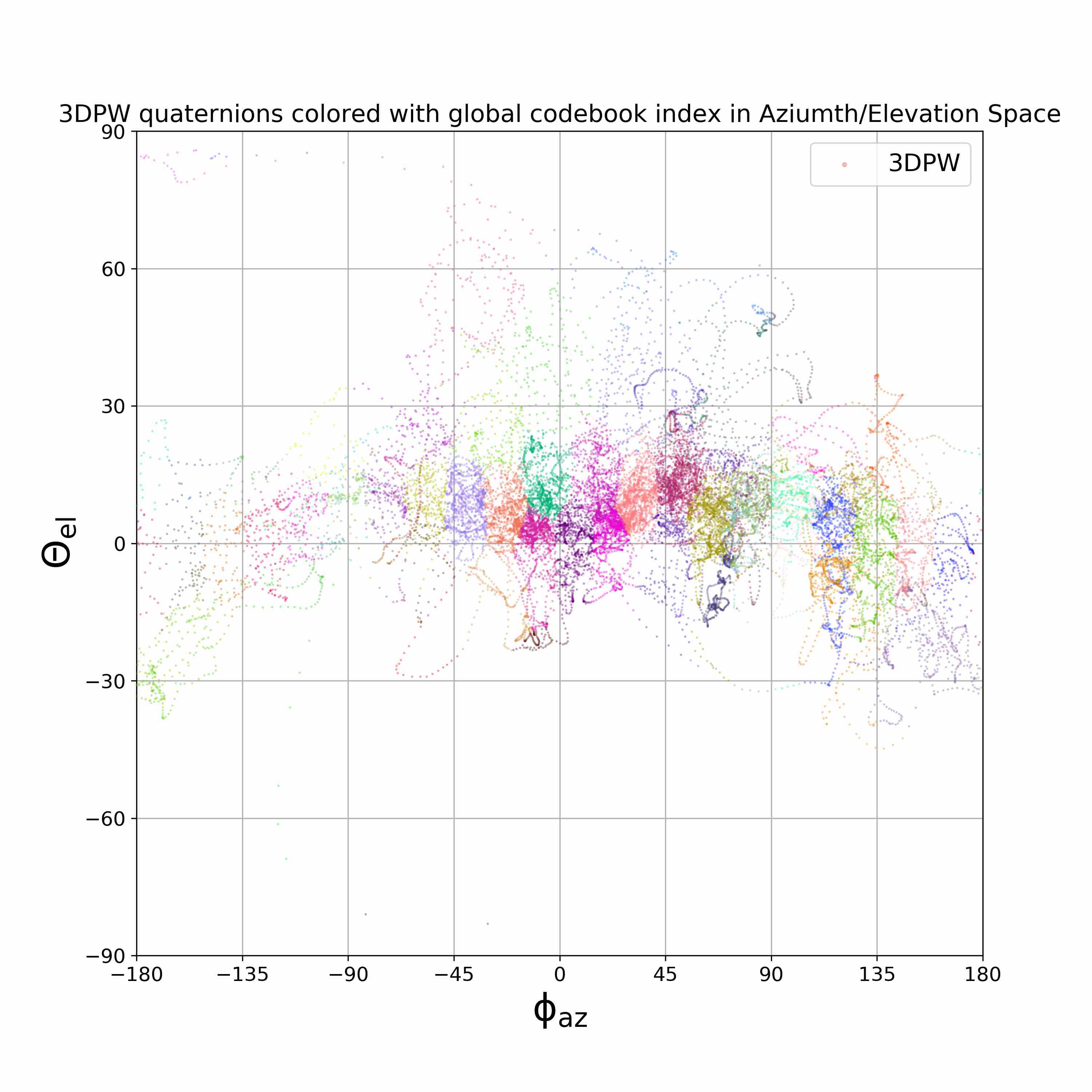}
\caption{\textsc{3DPW}} \label{fig:2e}
\end{subfigure}
\hspace*{\fill}
\begin{subfigure}{0.32\textwidth}
\includegraphics[width=\linewidth]{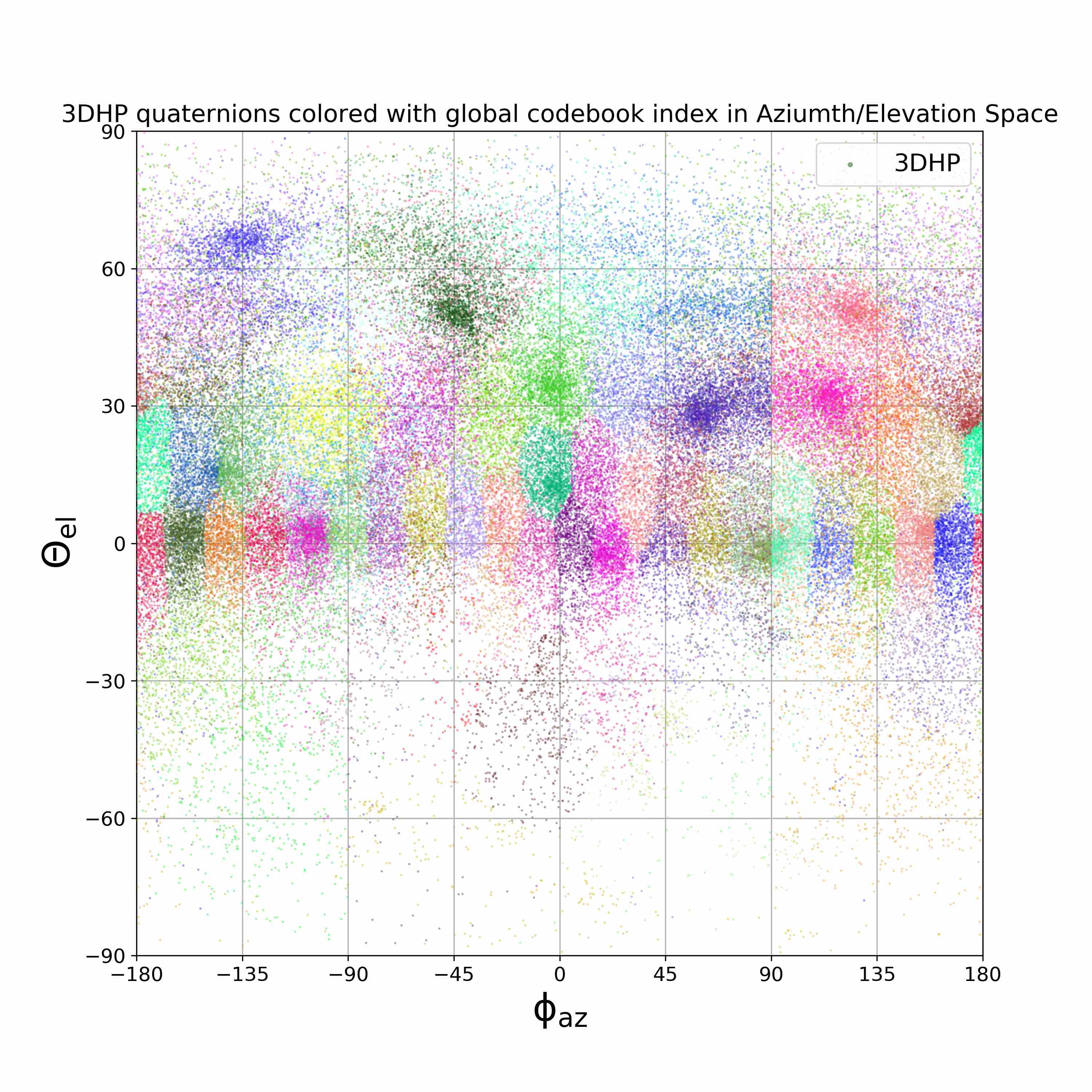}
\caption{\textsc{3DHP}} \label{fig:2f}
\end{subfigure}
\caption{\textbf{a}: Illustration of our body-centered coordinate frame (up
vector, right vector and front vector) relative to a camera-centered coordinate
frame. \textbf{b-f}: Camera viewpoint distribution 
of the 5 datasets color by quaternion cluster index. Quaternions (rotation 
between body-centered and camera frame) are
sampled from training sets and clustered using k-means. They are
also visualized in azimuth / elevation space following Fig
\ref{fig:view_direction}.}
\label{fig:body-centeredUFR}
\vspace{-0.15in}
\end{figure*}

\subsection{Predicting the camera viewpoint}

To predict the camera viewpoint relative to the body-centered coordinate frame
we considered three approaches: (i) direct regression of $q$, (ii)
quantizing the space or rotations and performing k-way classification, and
(iii) a combined approach of first predicting a quantized rotation
followed by regressing the residual from the cluster center. In our
experiments, we found that the classification-based loss yields less accurate
coordinate frame predictions 
but yielded the
largest improvements in the pose prediction branch (see Table \ref{tab:ablationstudy}).

To quantize the space of rotations, we use k-means to cluster the quaternions
into k=100 clusters. The clusters are computed from training data of a single
dataset (local clusters) or from all five datasets (global clusters).  We
visualize the global cluster centers in azimuth and elevation space in Fig
\ref{fig:body-centeredUFR} b-f, as well as randomly sampled quaternions from
H36M, GPA, SURREAL, 3DPW and 3DHP datasets. 

To regress the quaternion $q$ we simply add a branch to our base pose prediction
model consisting of a 1x1 convolutional layer to reduce the feature dimension to 4
followed by global average pooling and normalization to yield a unit 4-vector.
We train this variant using a standard squared-Euclidan loss on target $q^*$.
For classification, we use the same prediction $q$ but compute the probability
it belongs to the correct cluster using a softmax to get a distribution over
cluster assignments:
\[
p(c|q) = \frac{\exp(-\mu_c^T q)}{\sum_{i=1}^{k} \exp(-\mu_i^T q )}
\]
where $\{\mu_1,\mu_2,\ldots,\mu_k\}$ are the quaternions corresponding to cluster centers computed
by k-means.  We use the negative log-likelihood as the training loss,
\begin{equation*}
  {\mathcal{L}_q =   -log(p(c^*|q)) }
  \label{eqn:qclass}
\end{equation*}
where $c^*$ is the viewpoint bin that the training example was assigned
during clustering. Our final loss consists of both quaternion and pose terms:
${\mathcal{L} =  \lambda \mathcal{L}_q + \mathcal{L}_{\textit{pose}}}$.
 
\section{Experiments}
\label{sec:exp}


\paragraph{Data and evaluation metric.} To reduce the redundancy of the
training images  (30 fps video gives lots of duplicated images for network
training), we down sample 3DHP, SURREAL to 5 fps. Following
\cite{rootnet,Zhou_2017_ICCV}, we sample H36M to 10 fps, and use the protocol 2
(subject 1,3,5,7,8 for training and subject 9,11 for testing) for evaluation.
As GPA is designed as monocular image 3d human pose estimation, which is
already sampled, we follow \cite{gpa} and directly use the released set. Number
of images in train set and test set is shown in Table \ref{table:datasets}.
In addition,  we use the MPII dataset \cite{mpii}, a large scale in-the-wild
human pose dataset for training a more robust   pose model. It contains 25k
training images and 2,957 validation images. We use two metrics, first is mean
per joint position error (MPJPE), which is calculated between predicted pose
and ground truth pose.  The second one is PCK3D \cite{mono_3dhp2017}, which is the accuracy of joint prediction (threshold on MPJPE with 150mm).

 \begin{table}[t]
\begin{center}
{\scriptsize
\begin{tabular}{@{}ll c c c c c@{}}
\toprule
 & &\multicolumn{5}{c}{MPJPE (in mm, lower is better)} \\
 & Testing \textbackslash Training & H36M & GPA & SURREAL & 3DPW & 3DHP  \\
\hline
\multirow{5}{*}{Baseline} & H36M & \textbf{53.2} & 110.5 & 107.1 & 125.1  & 108.4\\
&GPA & 105.2 & \textbf{53.9} & 86.8 & 111.7 & 90.5 \\
&SURREAL & 118.6 & 103.2 & \textbf{37.2} & 120.8 & 108.2 \\
&3DPW & 108.7 & 116.4 & 114.2 & \textbf{100.6} & 113.3 \\
&3DHP & 111.8 & 123.9 & 120.3 & 139.7 &  \textbf{91.9} \\
\hline
\multirow{5}{*}{Our Method} & H36M & \textbf{52.0} & \textcolor{blue}{102.5} & 103.3 & 124.2 & \textcolor{blue}{95.6}  \\
& GPA & \textcolor{blue}{98.3} & \textbf{53.3} & 85.6 & 110.2 & 91.3  \\
& SURREAL & 114.0 & 101.2 & \textbf{37.1} & 113.8 & 107.2\\
& 3DPW & 109.5 & 112.0 & \textcolor{blue}{112.2} & \textcolor{blue}{\textbf{89.7}} & 105.9  \\
& 3DHP & 111.9 & 119.7 & 118.2 & 136.0 & \textbf{90.3} \\
\hline
\multicolumn{2}{l}{Same-Dataset Error Reduction $\downarrow$} & 1.2 & 0.6 & 0.1 & 10.9 & 1.5 \\
\multicolumn{2}{l}{Cross-Dataset Error Reduction $\downarrow$} & 10.6 & 18.6 & 9.1 & 13.1 & 20.4 \\
\bottomrule
\end{tabular}
}
\end{center}
\caption{Baseline cross-dataset test error and error reduction from the addition
of our proposed quaternion loss. Bold indicates the best performing model on 
each the test set (rows). Blue color indicates test set which saw greatest error
reduction. See appendix for corresponding tables of PCK and Procrustese 
aligned MPJPE.}
\label{table:baselineposenet}
\vspace{-0.15in}
\end{table}

\paragraph{Implementation Details.} As different datasets have diverse joint
configuration, we select a subset of 14 joints  that all datasets share to
eliminate the bias introduced by different number of joints during training.
We normalize the z value from ($-z_{\textit{max}}$,
$+z_{\textit{max}}$) to ($0,63$) for integral regression. $z_{\textit{max}}$
is 2400 mm based all 5 set. We use PyTorch to implement our network. The
ResNet-50 \cite{resnet} backbone is initialized using the pre-trained weights
on the ImageNet dataset. We use the Adam \cite{adam} optimizer with a
mini-batch size of 128. The initial learning rate is set to 1 $\times$
$10^{-3}$ and reduced by a factor of 10 at the 17th epoch, we train 25 epochs
for each of the dataset. We use 256 $\times$ 256 as the size of the input image
of our network. We perform data augmentation including rotation, horizontal
flip, color jittering and synthetic occlusion following \cite{rootnet}. We set
$\lambda$ to 0.5 for the quaternion loss which is validated on 3DPW validation
set.



\subsection{Cross-dataset evaluation} 

We list the cross-dataset
baseline and our improved  results in Table \ref{table:baselineposenet}. The
bold numbers indicate the best performing model on the test set. As expected,
the best performance occurs when the model is trained and evaluated on the same
set. The numbers marked with blue color indicate the test set where the error
reduction is most significant, using our proposed quaternion loss.



\paragraph{Training on H36M.} Adding the quaternion loss reduces total cross-dataset
error by 10.6 mm (MPJPE), while the same-dataset error reduction is 1.2 mm (MPJPE).
This may be explained by the error on H36M already being low. The largest error reduction is on GPA  (6.9 mm) which we attribute to de-biasing the azimuth distribution difference as shown in Fig~\ref{fig:az}. 


\paragraph{Training on GPA.} The total cross-dataset error reduction is 18.6 mm
(MPJPE), and the same data error reduction is 0.6 mm (MPJPE). We attribute this to
the bias during capture \cite{gpa}: the coverage of camera viewing directions
is centered in the range of $-60$ to 90 degrees azimuth (as in Fig
\ref{fig:az}). The largest cross-data set error reduction occurs for H36M, with 8.0
mm. This further demonstrates that the view direction distribution is largely
different from H36M. 



\begin{figure}[t]
\begin{center}
\begin{tabular}{cc}
\includegraphics[width=0.48\linewidth]{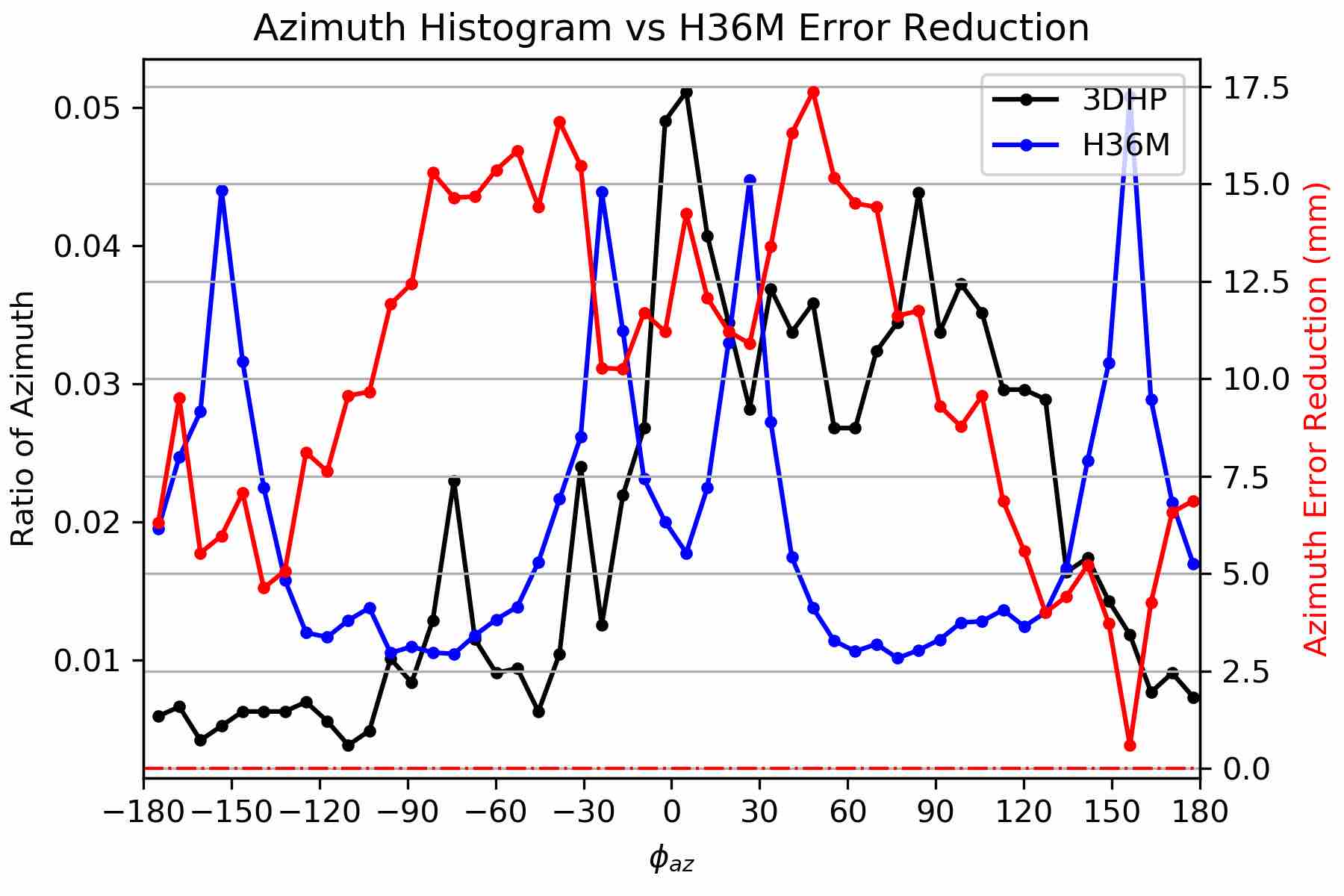} &
\includegraphics[width=0.48\linewidth]{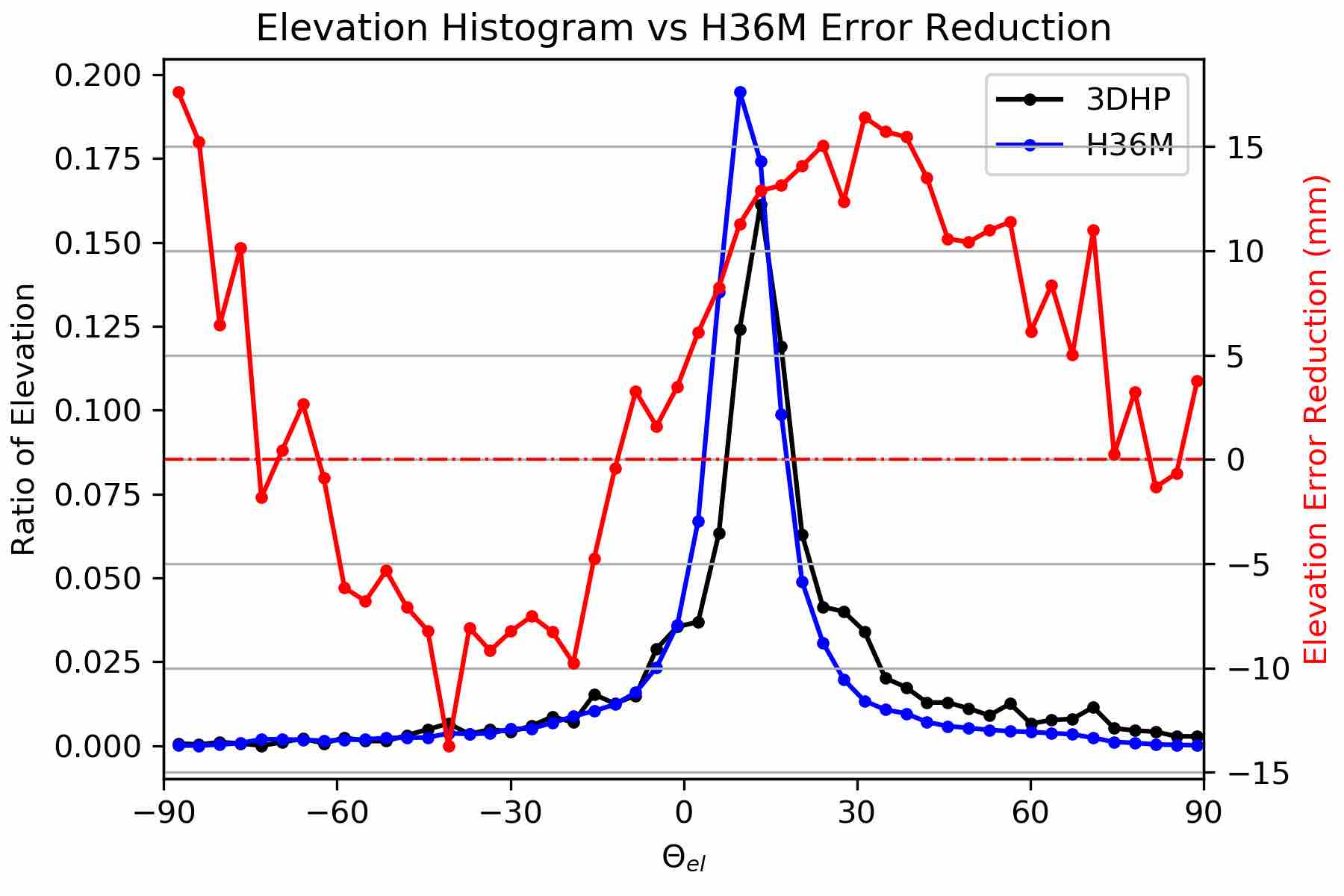}
\end{tabular}
\end{center}
\caption{We visualize viewpoint distributions for train (3DHP) and test (\textcolor{blue}{H36M}) 
overlayed with the \textcolor{red}{reduction} in pose prediction error relative to baseline}
\label{fig:erroreduction}
\vspace{-0.15in}
\end{figure}

\paragraph{Training on SURREAL.} Adding the quaternion loss reduces the
cross-dataset error by 9.1 mm (MPJPE), while the same-dataset error reduction is 0.1
mm (MPJPE).  We attribute this to the fact that viewpoint distribution on
SURREAL itself is already uniform as in Fig \ref{fig:az}. We can see distribution
over azimuths is quite uniform. Thus adding more
supervision in the form of quaternion loss helps little. The most error
reduction (2.0mm) is observed on 3DPW. We attribute this
to the fact that 3DPW is strongly biased dataset in terms of view direction,
and the quaternion loss helps reduce the view difference between SURREAL and
3DPW. 


\begin{table}[t]
\begin{center}
{\scriptsize
\begin{tabular}{@{}l c c c c c@{}}
\toprule
 & \multicolumn{5}{c}{MPJPE (in mm, lower is better)}  \\
Metric \textbackslash Training Set & H36M & GPA & SURREAL & 3DPW & 3DHP \\
\hline
Same-Dataset Error Reduction $\downarrow$ & 0.6 & 4.2        & 0.2 & 7.6 & 1.2 \\
Cross-Dataset Error Reduction  $\downarrow$ & 2.4 & 12.3 & 1.9 & 10.1 &  9.3 \\
\bottomrule
\end{tabular}}
\end{center}
\caption{Retraining the model of Zhou~\etal~\cite{Zhou_2017_ICCV} using our viewpoint prediction
loss yields also shows significant decrease in prediction error, demonstrating the generality of
our finding. See appendix for full table of numerical results.}
\label{tab:secondmethod}
\vspace{-0.15in}
\end{table}

\begin{table}[t]
\begin{center}
\centering
{\scriptsize
\begin{tabular}{ c c c c c c c}
\toprule
Datasets & Baseline & C & R & C+R & C+local cluster & C+cannonical pose      \\
\hline
3DPW (MPJPE (mm)) & 100.6 & 89.7 & 94.0 & 93.2 & 93.1 & 100.3   \\
\bottomrule
\end{tabular}
}
\end{center}
\caption{Ablation analysis: we compare the performance of our proposed camera
view-point loss using classification (C), regression (R), using both (C+R);
using per-dataset clusterings (local) rather than the global clustering; and
adding a third branch which also predicts pose in canonical body-centered
coordinates.}
\label{tab:ablationstudy}
\vspace{-0.15in}
\end{table}

\paragraph{Training on 3DPW.} The error is reduced by 10.9 mm (MPJPE) on itself
(also the most error reduction one with model trained on 3DPW), and the cross-dataset
error reduction is 13.1 mm (MPJPE). From the Fig \ref{fig:az} we can see, in
terms of azimuth, 3DPW has a strong bias towards $-30$ degree to 60 degree. As
during capture, the subject is always facing towards the camera to make it
easier for association between  the subject (there are multiply persons in
crowded scene) and IMU sensors, this bias seems inevitable and quaternion loss
is helpful for this kind of in the wild dataset to reduce view direction bias.
It is also verified in 3DHP, where half of the test set is in the wild, and
have view direction bias. 


\paragraph{Training on 3DHP.}   Adding the quaternion loss reduces the total 
cross-dataset error by 20.4 mm, while the same-dataset error reduction is 1.5 mm (MPJPE).
During the capture, 3DHP capture images from a wide range of viewpoints. We can see from the Fig \ref{fig:view_direction} that the azimuth
of 3DHP is the most uniformly distributed of the real datasets. Thus treating
it as training set will enable the network to be robust to view direction. 
We also calculate error reduction conditioned on azimuth and
elevation on the H36M test set (Fig \ref{fig:erroreduction}). The blue/black line is azimuth and elevation histogram
distribution for H36M/3DHP training sets while the red line shows relative error
reduction for H36M. We can see the error is reduced more where H36M has
fewer views relative to 3DHP. 


\subsection{Effect of Model Architecture and Loss Functions}
To demonstrate the generalization of
our approach to other models, we also added a viewpoint prediction branch
to the model of \cite{Zhou_2017_ICCV} which utilizes a different model 
architecture. We observe similar results in terms of improved generalization
(see Table \ref{tab:secondmethod} and appendix). We note that while our
primary baseline model \cite{rootnet} uses camera intrinsic parameters to
back-project, \cite{Zhou_2017_ICCV} utilizes an average bone-length estimate
from the training set which results in higher prediction errors across datasets.


\begin{table}[!t]
\centering
{\scriptsize
\begin{tabular}{@{}l c c c c c|c c c c c@{}}
\Xhline{2\arrayrulewidth}
\toprule
 &\multicolumn{5}{c}{MPJPE$\downarrow$: lower is better} & \multicolumn{5}{c}{PCK3D$\uparrow$: higher is better} \\
& H36M & GPA & SURREAL & 3DPW & 3DHP & H36M & GPA & SURREAL & 3DPW & 3DHP\\
\hline
Mehta \cite{mono_3dhp2017} & 72.9 & - & - & - & -                
& - & - & - & - & 64.7  \\
Zhou \cite{Zhou_2017_ICCV} & 64.9 & \underline{96.5} & - & - & - 
& - & \underline{82.9} & - & - & 72.5   \\
Arnab\cite{temporalinthewild} & 77.8 & - & - & - & -             
& - & - & - & - & -   \\
Kanazawa \cite{HMR} & 88.0 & - & - & - & 124.2                   
& - & - & - & - & 72.9   \\
Kanazawa \cite{HMMR} & - & - & - & \underline{127.1} & -         
& - & - & - & $\mathbf{86.4^*}$ & - \\
Moon \cite{rootnet} & 54.3 & - & - & - & -                       
& - & - & - & - & -    \\
Kolotouros \cite{GraphCMR}  & 78.0 & - & - & - & -               
& - & - & - & - & -   \\
Tung\cite{selfmocap} &  98.4 &  -& $64.4^*$ &  -& - 
& - & - & - & - & -   \\
Varol\cite{bodynet} & $\mathbf{51.6^*}$ & - & \underline{49.1} & - & -   
& - & - & - & - & -   \\
Habibie \cite{inthewildintermediate} & 65.7 & - & - & - & \underline{91.0}  
& - & - & - & - & \underline{82.0}  \\
Yu \cite{skeletonrepresentation} & 59.1 &  - & - & - & -         
& - & - & - & - & -   \\
Ours & \underline{52.0} & \textbf{53.3} & \textbf{37.1} & \textbf{89.7} & \textbf{90.3}
& \textbf{96.0} & \textbf{96.8} &  \textbf{97.3} &  \underline{84.6} & \textbf{84.3} \\
\bottomrule
\end{tabular}
}
\caption{Comparison to state-of-the-art performance. 
There are many missing entries, indicating how infrequent it is to perform
multi-dataset evaluation.  Our model provides a new state-of-the art baseline
across all 5 datasets and can serve as a reference for future work.
* denotes training using extra data or annotations (e.g. segmentation). 
Underline denotes the second best results.
}
\label{table:stateofart}
\end{table}

\begin{figure}[t]\centering
   \includegraphics[width=1\linewidth]{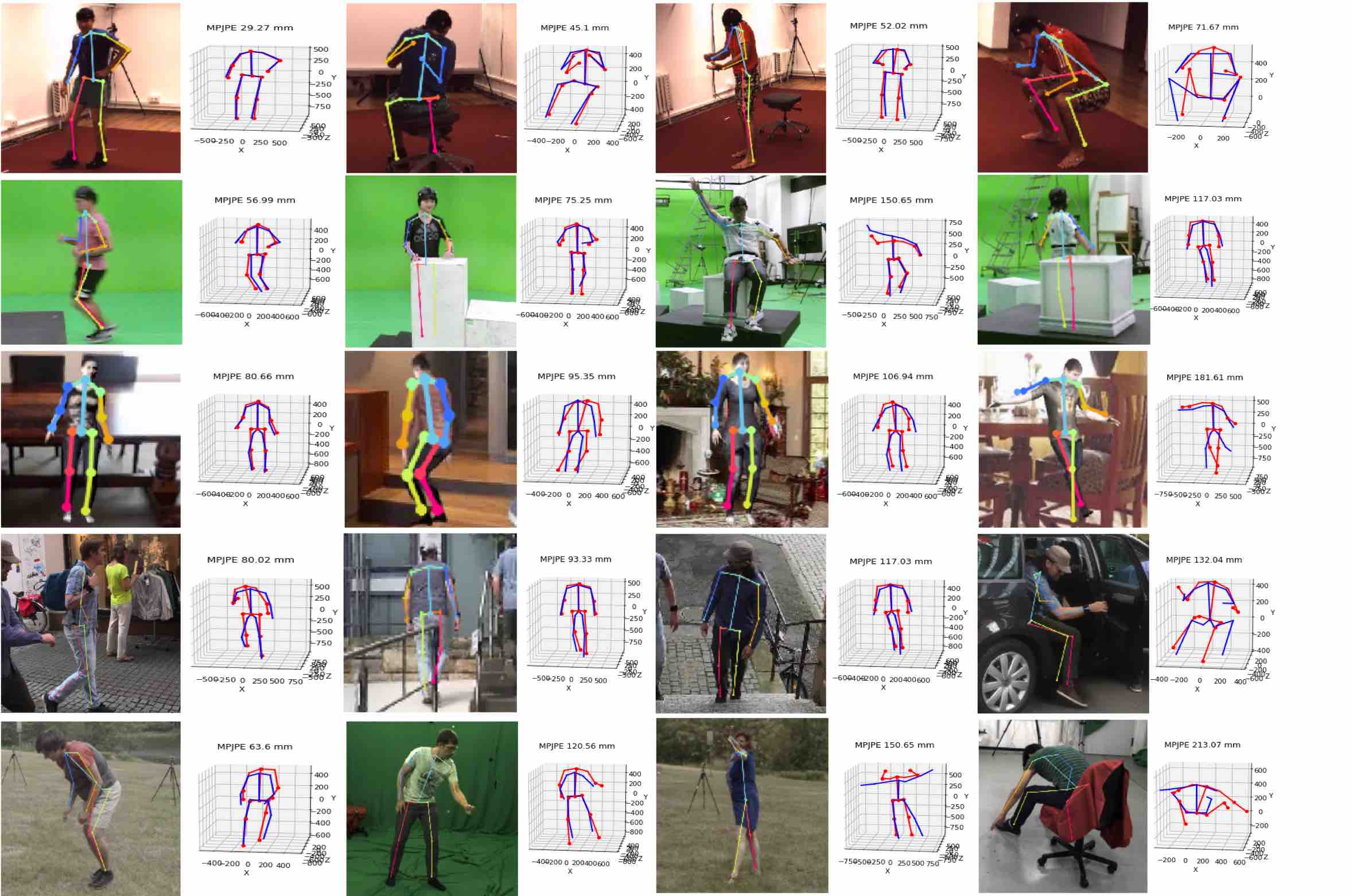}
   \caption{Model predictiosn on 5 datasets from model trained on Human3.6M dataset.
   The 2d joints are overlaid with the original image, while the
   \textcolor{red}{3d prediction (red)} is overlaid with \textcolor{blue}{3d
   ground truth (blue)}. 3D prediction is \textbf{visualized in body-centered
   coordinate} rotated by the relative rotation between ground truth
   camera-centered coordinate and body-centered coordinate. From top to bottom
   are \textsc{H36M, GPA, SURREAL, 3DPW} and \textsc{3DHP} datasets. We rank
   the images from left to right in order of increasing MPJPE.} 
\label{fig:qualitativeresults}
\vspace{-0.15in}
\end{figure}

\paragraph{Ablation study} To explore whether our methods are robust to
different k-means initialization, we repeat k-means 4 times and report
performance on 3DPW. We find the range of the MPJPE is within  $90\pm 0.4$
([89.9, 89.6, 90.2, 89.7]) mm. We also vary the number of clusters to select
the best $k  \in \{10,24,50,100,200,500\}$, with corresponding errors [93.0,
95.2, 92.3, 89.7, 93.0,93.2]. We find k=100 is the best number with at most 6
mm reduction compared to k=24.  In Table \ref{tab:ablationstudy}, the error
of global clusters is 3.4 mm error less than local, per-dataset clusters,
demonstrating training on global clusters is better than local clusters
which are biased towards the training set view distribution.  In terms of
choice for quaternion regression, k-way classification reduced error by 4.3 mm
compared to regression. While utilizing both classification and regression
losses gives error than regression only.


Finally, we also consider adding a third branch and loss function to the model
which also predicts the 3D pose in the body-centered coordinate system. This is
related to the hand pose model of~\cite{zb2017hand}, although we don't use this
prediction of canonical pose at test time. This variant performs global pooling
on the ResNet feature map after upsampling followed by a two layer MLP that
predicts the viewpoint $q$ and canonical pose. When training with this additional
branch we find the camera-centered pose predictions show no improvement over
baseline (Table \ref{tab:ablationstudy}). We also observe that the canonical
pose predictions have higher error than the camera-centered predictions which
is natural since the the model can't directly exploit the direct correspondence
between the 2D keypoint locations and the 3D joint locations. 


\subsection{Comparison with state-of-the-art performance} 

Table \ref{table:stateofart} compares the proposed approach with the state-of-the-art
performance on all 5 datasets. Note that our method is the first to evaluate 3d
human pose estimation on the five representative datasets reporting both MPJPE
and PCK3D, which fills in some blanks and serves as a useful baseline for future 
work. As can be seen, our method achieves state-of-the-art performance on
H36M/GPA/SURREAL/3DPW/3DHP datasets in terms of MPJPE. While
\cite{HMMR} uses additional data (both H36M and 3DHP, and LSP
together with MPII) to train, they have slightly better performance on 3DHP in
terms of PCK3D. 

\paragraph{Qualitative Results:} We visualize the prediction on the 5 datasets
with model trained on H36M using our proposed method in Fig
\ref{fig:qualitativeresults}. The 2d joint prediction is overlaid with cropped
images while the 3d joint prediction is visualized in our proposed
body-centered coordinates. From top to bottom are H36M, GPA, SURREAL, 3DPW and
3DHP datasets. We display the images from left to right in ascending order by
MPJPE.

\section{Conclusions}
\label{sec:conclusion}
In this paper, we observe strong dataset-specific biases present in the
distribution of cameras relative to the human body and propose the use of
body-centered coordinate frames. Utilizing the relative rotation between
body-centered coordinates and camera-centered coordinates as an additional
supervisory signal, we significantly reduce the 3d joint prediction error and improve generalization in cross-dataset 3d human pose evaluation. Out 
model also achieves state-of-the-art performance on all same-dataset evaluations. We hope that our cross-dataset analysis is useful for future work and serves as a resource to guide future dataset collection.

\section{Acknowledgement}
Acknowledgements:  This work was supported in part by NSF grants IIS-1813785, IIS-1618806, 
and a hardware gift from NVIDIA.

\bibliographystyle{splncs04}
\bibliography{egbib}

\newpage

\appendix
\section*{Appendix}

 In the appendix, we \textbf{(1.)} visualize the UMAP embedding \cite{mcinnes2018umap-software} of view-dependent pose (root-relate coordinates) from the five datasets. \textbf{(2.)} We provide results for other evaluation metrics (PMPJPE and PCK3D) that parallel the MPJPE results shown in the main paper.   We also provide more detailed results showing the effectiveness of our quaternion loss in improving generalization of an alternate model of Zhou~\cite{Zhou_2017_ICCV}.  \textbf{(3.)} We visualize the distribution of viewpoints of five datasets in azimuth and elevation with cluster centers overlaid, \textbf{(4.)} We show selected examples based on the quaternion distribution pattern from five datasets. \textbf{(5.)} Finally, we show qualitative comparisons of training on each single dataset and testing across the five datasets, and training on five different datasets while testing on the same image from single dataset.

\section{UMAP Visualization}
\begin{figure*}[t]
\centering
\begin{subfigure}{0.32\textwidth}
\includegraphics[width=\linewidth]{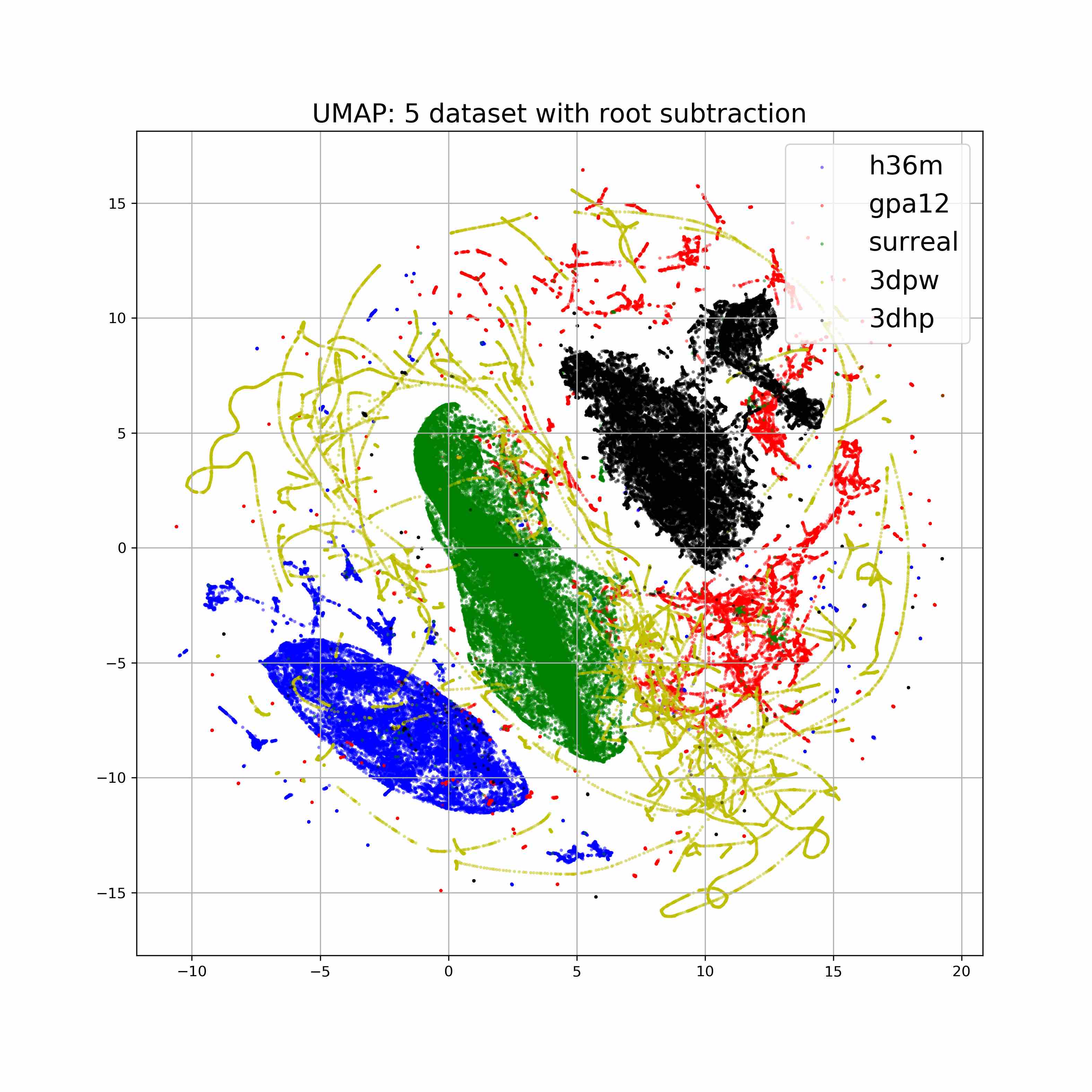}
\caption{\textsc{UMAP with only root-subtraction}} \label{fig:ua}
\end{subfigure}
\hspace*{\fill}
\begin{subfigure}{0.32\textwidth}
\includegraphics[width=\linewidth]{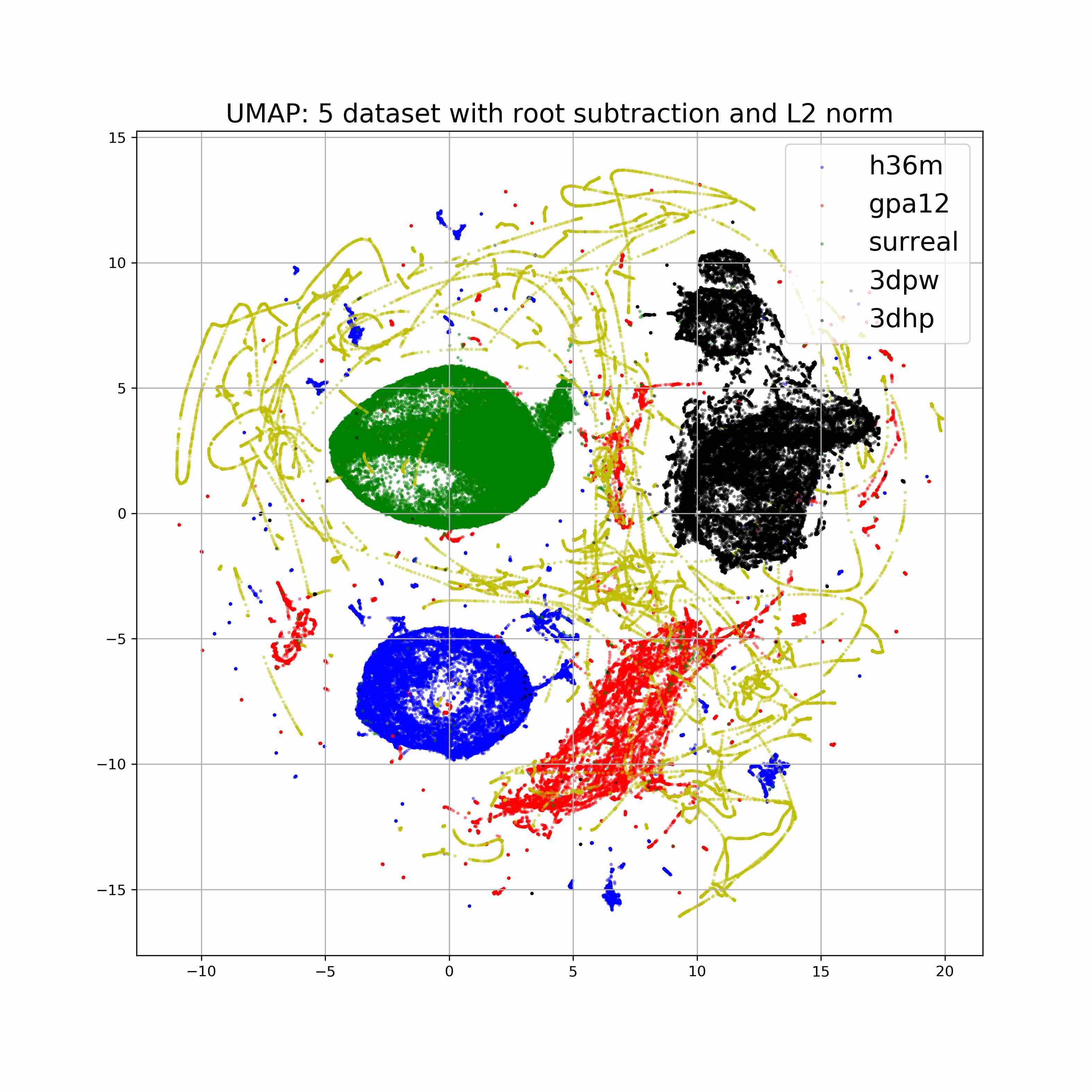}
\caption{\textsc{UMAP with root-subtraction and L2 normalization}} \label{fig:ub}
\end{subfigure}
\hspace*{\fill}
\begin{subfigure}{0.32\textwidth}
\includegraphics[width=\linewidth]{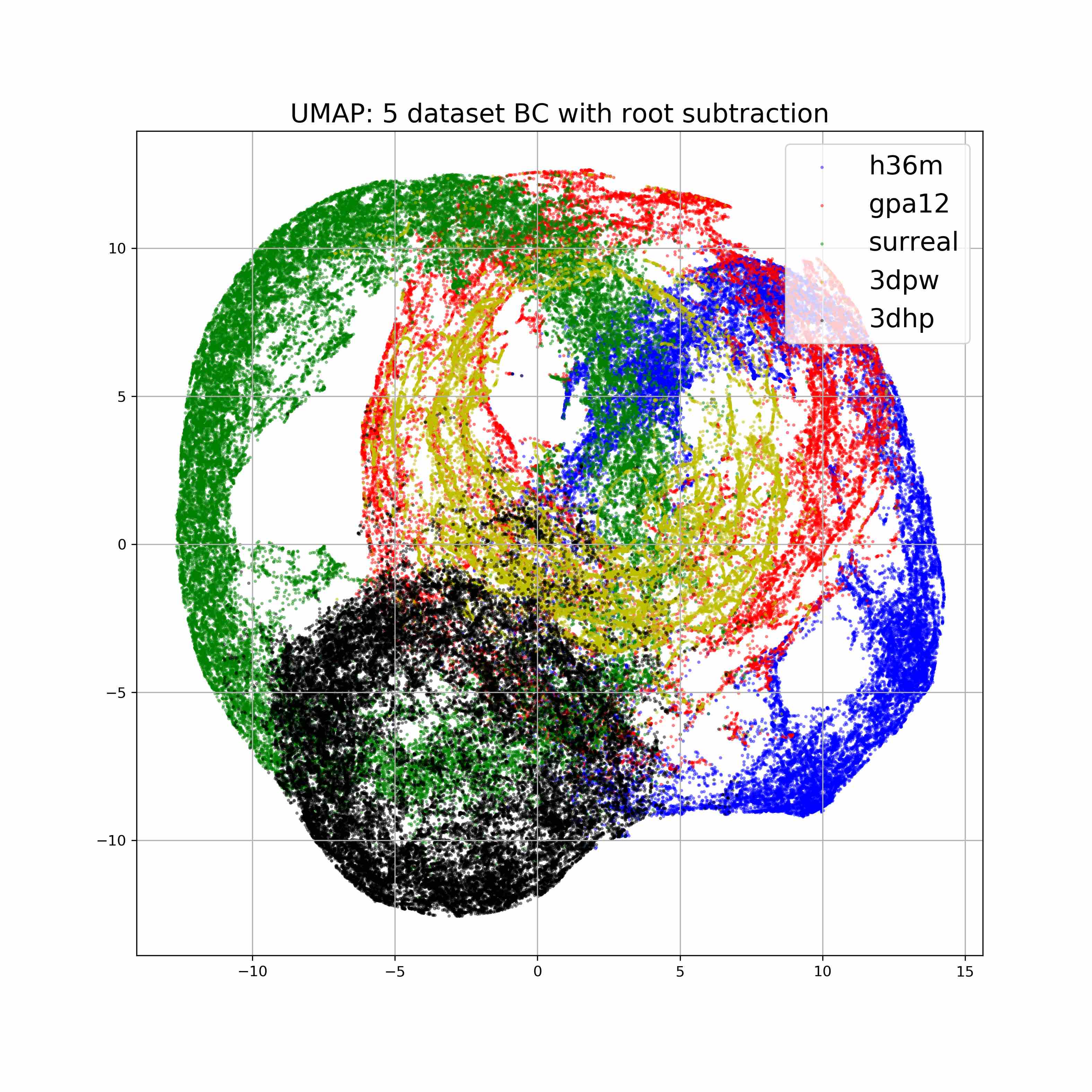}
\caption{\textsc{UMAP body-centered coordinates with only root-subtraction}} \label{fig:uc}
\end{subfigure}
\caption{\small Distribution of view-dependent, view-independent body-centered pose, visualized 
   as a 2D embedding produced with UMAP \cite{mcinnes2018umap-software}.}
\label{fig:umapfigures}
\end{figure*}

We visualize the UMAP \cite{mcinnes2018umap-software} embedding of view-dependent coordinate (root-relate coordinate) of H36M~\cite{h36m_pami}, GPA~\cite{gpa} , SURREAL~\cite{varol17_surreal}, 3DPW~\cite{inthewildeccv2018} and 3DHP~\cite{mono_3dhp2017} datasets in Fig \ref{fig:ua}.
We further normalize out skeleton size and visualize in Fig \ref{fig:ub}. To compare with view-independent coordinate (body-center coordinate), we visualize them before L2 normalization in Fig \ref{fig:uc}. We can see the body-centered, size normalized pose distribution (main paper) shows much higher overlap across datasets while the root-relative coordinates implicitly which encode camera orientation provide distinguishable information (dataset bias).

\section{PMPJPE, PCK3D results on posenet \cite{rootnet} and MPJPE results on Zhou~\cite{Zhou_2017_ICCV}}
 We provide  PMPJPE in Table \ref{table:pampjpe} and PCK3d in Table \ref{table:pck3d} to demonstrate the effectiveness of adding quaternion loss to PoseNet \cite{rootnet}. To demonstrate the utility of our quaternion loss on other models, we also show results based on retraining the model of \cite{Zhou_2017_ICCV} in Table \ref{table:mpjpesecond} with MPJPE metric.

\begin{table}[t]
\begin{center}
{\scriptsize
\begin{tabular}{@{}ll c c c c c@{}}
\toprule
 & &\multicolumn{5}{c}{PA-MPJPE (in mm, lower is better)} \\
 & Testing \textbackslash Training & H36M & GPA & SURREAL & 3DPW & 3DHP  \\
\hline
\multirow{5}{*}{Baseline} & H36M & \textbf{43.4} & 75.0 & 69.6 & 91.3  & 75.0\\
&GPA & 75.4 & \textbf{41.7} & 66.3 & 84.4 & 70.2 \\
&SURREAL & 76.5 & 73.5 & \textbf{31.8} & 85.8 & 77.9 \\
&3DPW & 68.0 & 66.9 & 64.3 & \textbf{68.7} & 68.1 \\
&3DHP & 88.5 & 91.2 & 86.9 & 111.3 &  \textbf{71.4} \\
\hline
\multirow{5}{*}{Our Method} & H36M & \textbf{42.5} & \textcolor{blue}{69.5} & 67.5 & 91.4 & \textcolor{blue}{72.6}  \\
& GPA & \textcolor{blue}{71.4} & \textbf{40.9} & 65.6 & 81.4 & 70.6  \\
& SURREAL & 75.9 & 71.7 & \textbf{31.7} & \textcolor{blue}{82.1} & 76.9\\
& 3DPW & 68.3 & 65.1 & 63.8 & \textbf{65.2} & 66.4  \\
& 3DHP & 89.0 & 89.7 & \textcolor{blue}{85.9} & 109.2 & \textbf{70.6} \\
\hline
\multicolumn{2}{l}{Same-Dataset Error Reduction $\downarrow$} & 0.9 & 0.8 & 0.1 & 3.2 & 0.8 \\
\multicolumn{2}{l}{Cross-data Error Reduction $\downarrow$} & 2.9 & 10.6 & 4.3 & 8.7 & 4.7\\

\bottomrule
\end{tabular}
}
\end{center}
\caption{Baseline cross-dataset test error and error reduction  (Procrustese 
aligned MPJPE) from the addition
of our proposed quaternion loss. Bold indicates the best performing model on 
each the test sets (rows). Blue color indicates test set which saw greatest error
reduction.}
\label{table:pampjpe}
\vspace{-0.15in}
\end{table}

\begin{table}[t]
\begin{center}
{\scriptsize
\begin{tabular}{@{}ll c c c c c@{}}
\toprule
 & &\multicolumn{5}{c}{PCK3D (accuracy, higher is better)} \\
 & Testing \textbackslash Training & H36M & GPA & SURREAL & 3DPW & 3DHP  \\
\hline
\multirow{5}{*}{Baseline} & H36M & \textbf{95.7} & 75.7 & 52.3  & 70.6  & 77.8\\
&GPA & 78.3 & \textbf{96.3} & 58.8 & 76.2  & 84.5 \\
&SURREAL &  76.4 & 84.5 & \textbf{97.2}  & 73.6  & 81.0  \\
&3DPW & 83.2 & 78.7 & 54.5  & \textbf{82.1} & 81.7 \\
&3DHP & 76.1 & 70.3 &  44.8 & 68.4 &  \textbf{84.2} \\
\hline
\multirow{5}{*}{Our Method} & H36M & \textbf{96.0} & \textcolor{blue}{78.9} & 52.6 & 72.8 & \textcolor{blue}{78.3}  \\
& GPA & \textcolor{blue}{81.5} & \textbf{96.8} & \textcolor{blue}{59.3} & 76.4 & 84.8  \\
& SURREAL & 80.0 & 84.8 & \textbf{97.3} & \textcolor{blue}{76.2} & 81.3 \\
& 3DPW & 83.2 & 80.8 & 54.7 & \textbf{84.6} & 81.7  \\
& 3DHP & 76.1 & 73.5 & 45.1 & 70.3 & \textbf{84.3} \\
\hline
\multicolumn{2}{l}{Same-Dataset Accuracy Increase $\uparrow$} & 0.3  & 0.5  & 0.1  & 2.5  & 0.1  \\
\multicolumn{2}{l}{Cross-data Accuracy Increase $\uparrow$} & 6.8  & 8.8 & 1.3  & 6.9  & 1.1 \\

\bottomrule
\end{tabular}
}

\end{center}
\caption{Baseline cross-dataset test accuracy and accuracy increases  (PCK3D) from the addition
of our proposed quaternion loss. Bold indicates the best performing model on 
each the test set (rows). Blue color indicates test set which saw greatest accuracy 
increase.}
\label{table:pck3d}
\vspace{-0.15in}
\end{table}

 

\begin{table}[t]
\begin{center}
{\scriptsize
\begin{tabular}{@{}ll c c c c c@{}}
\toprule
 & &\multicolumn{5}{c}{MPJPE (in mm, lower is better)} \\
 & Testing \textbackslash Training & H36M & GPA & SURREAL & 3DPW & 3DHP  \\
\hline
\multirow{5}{*}{Baseline} & H36M & \textbf{72.5} & 126.0 & 116.6 & 135.5  & 118.0\\
&GPA & 110.5 & \textbf{76.6} & 97.3 & 116.2 & 100.6 \\
&SURREAL & 129.6 & 116.0 & \textbf{54.1} & 132.3 & 118.7 \\
&3DPW & 120.1 & 121.9 & 120.2 & \textbf{108.5} & 119.8 \\
&3DHP & 122.9 & 133.6 & 128.5 & 148.0 &  \textbf{104.5} \\
\hline
\multirow{5}{*}{Our Method} & H36M & \textbf{71.9} & \textcolor{blue}{122.2} & \textcolor{blue}{115.4} & 134.4 & 109.9  \\
& GPA & \textcolor{blue}{109.9} & \textbf{72.4} & 97.8 & 115.3 & \textcolor{blue}{102.0}  \\
& SURREAL & 129.2 & 113.5 & \textbf{53.9} & \textcolor{blue}{126.5} & 119.4\\
& 3DPW & 119.1 & 119.3 & 119.9 & \textbf{100.9} & 116.5  \\
& 3DHP & 122.5 & 130.2 & 127.6 & 145.7 & \textbf{103.3} \\
\hline
\multicolumn{2}{l}{Same-Dataset Error Reduction $\downarrow$} & 0.6 & 4.2 & 0.2 & 7.6 & 1.2 \\
\multicolumn{2}{l}{Cross-data Error Reduction $\downarrow$} & 2.4 & 12.3 & 1.9 & 10.1 & 9.3\\

\bottomrule
\end{tabular}
}
\end{center}
\caption{Retraining the model of Zhou~\etal~\cite{Zhou_2017_ICCV} using our viewpoint prediction
loss also shows significant decrease in prediction error, demonstrating the generality of
our finding.}
\label{table:mpjpesecond}
\vspace{-0.15in}
\end{table}

\section{Quaternion and cluster centers}
 Instead of colorizing each quaternion with cluster index, we directly visualize quaternion with the same color within each dataset in Fig~\ref{fig:body-centeredUFR}, and also plot the cluster centers in the azimuth and elevation space.

\begin{figure*}[t]
\centering
\begin{subfigure}{0.36\textwidth}
\includegraphics[width=\linewidth]{anatomyC.jpg}
\caption{Body-centered coordinate} \label{fig:2a}
\end{subfigure}
\hspace*{\fill}
\begin{subfigure}{0.30\textwidth}
\includegraphics[width=\linewidth]{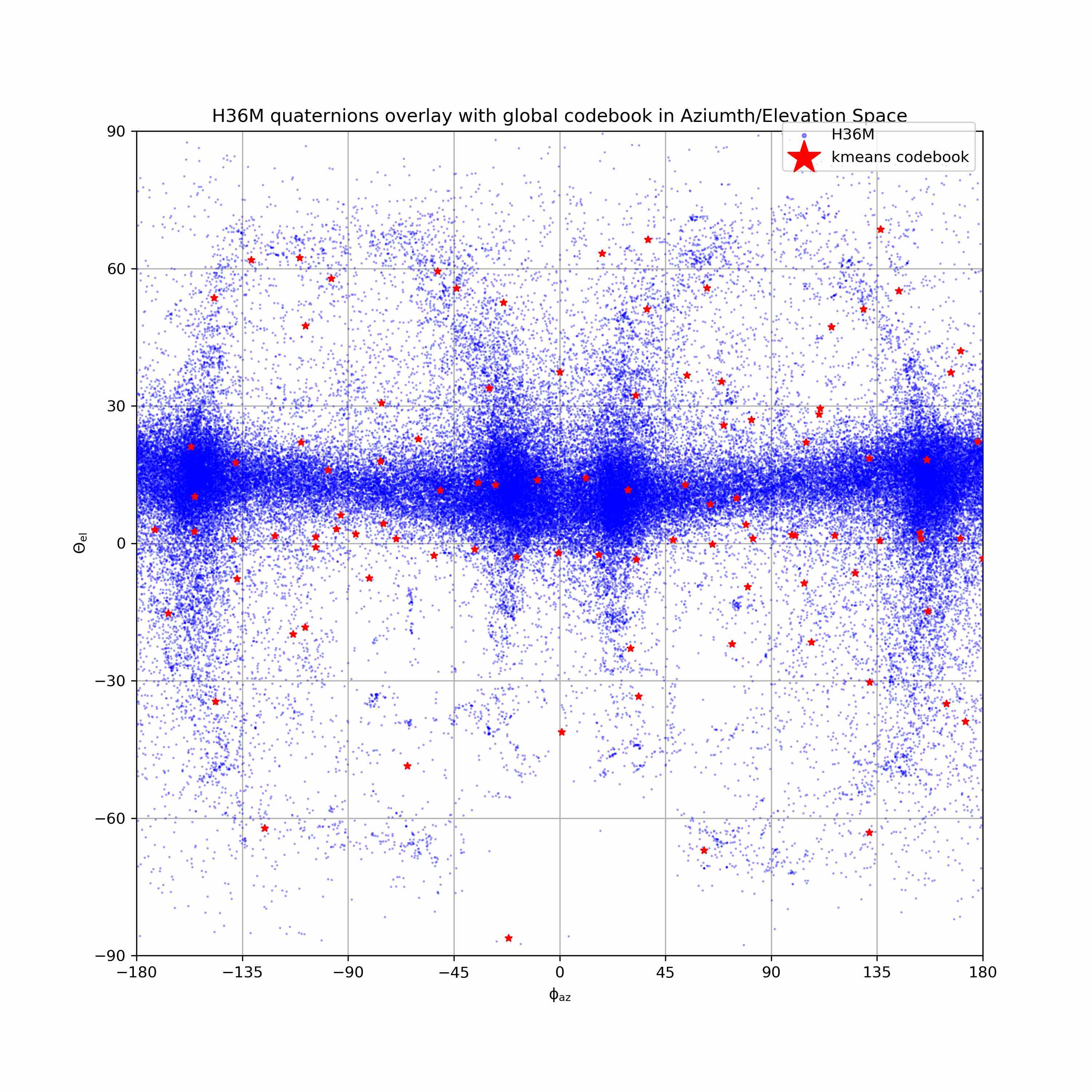}
\caption{\textsc{H36M}} \label{fig:2b}
\end{subfigure}
\hspace*{\fill}
\begin{subfigure}{0.30\textwidth}
\includegraphics[width=\linewidth]{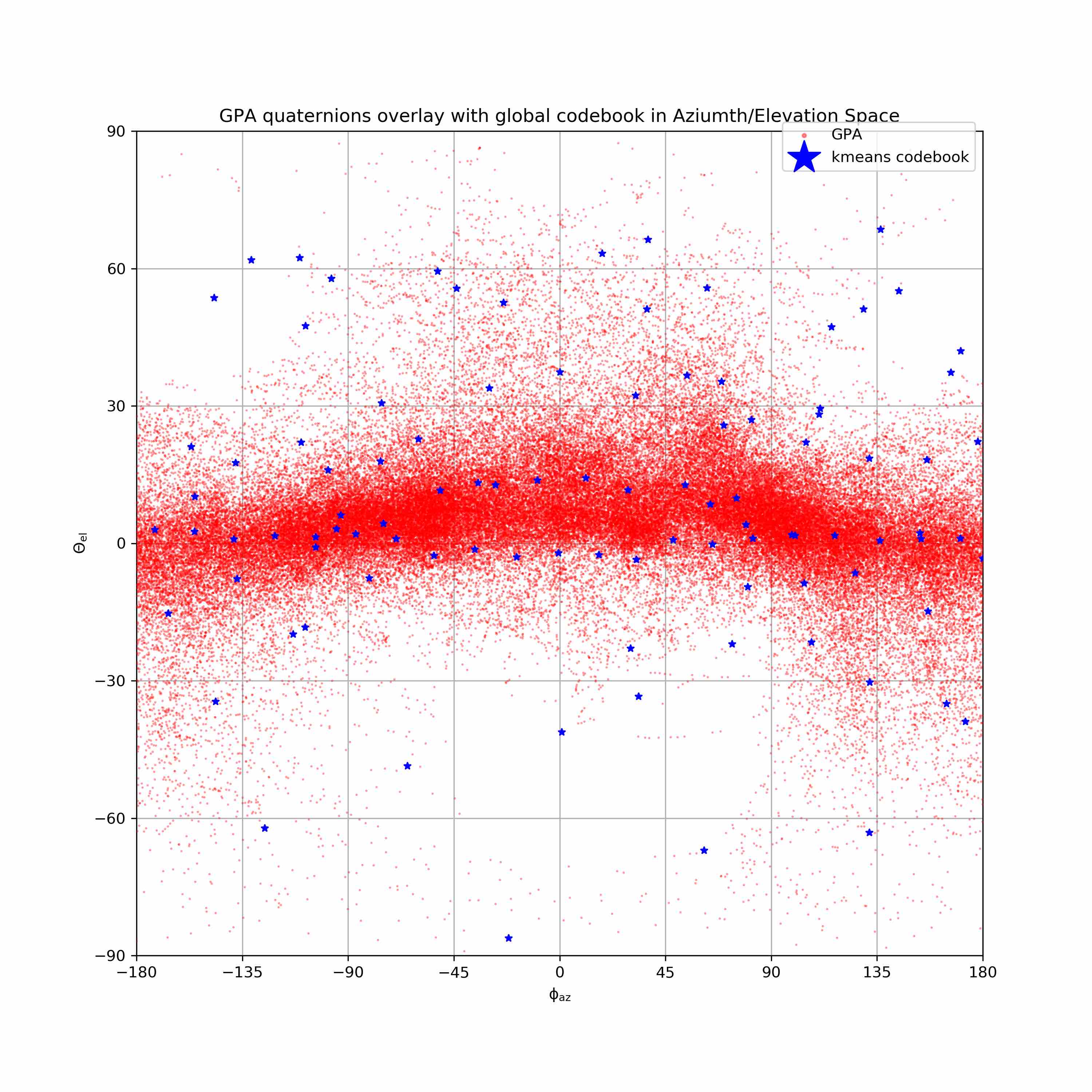}
\caption{\textsc{GPA}} \label{fig:2c}
\end{subfigure}
\hspace*{\fill}
\begin{subfigure}{0.32\textwidth}
\includegraphics[width=\linewidth]{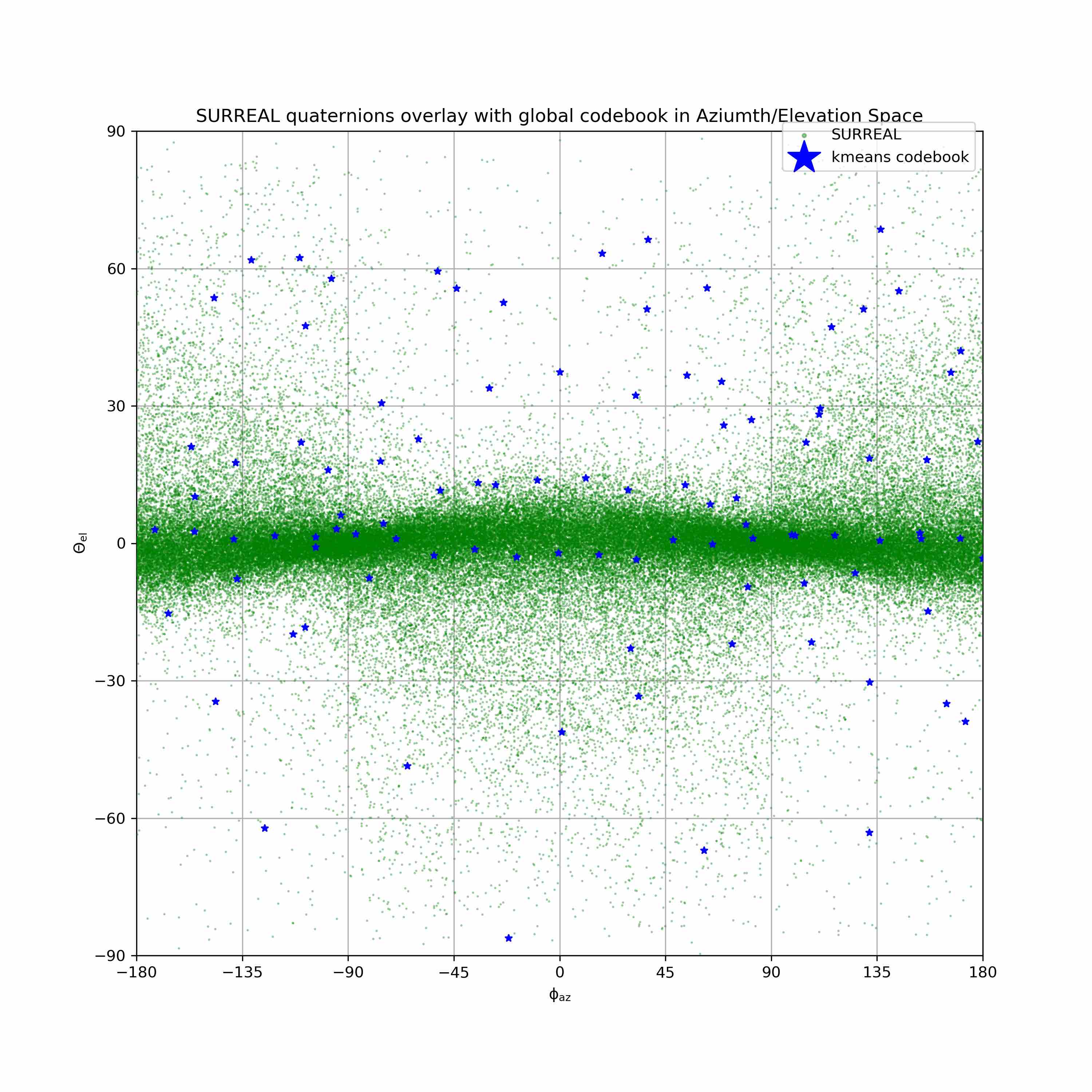}
\caption{\textsc{SURREAL}} \label{fig:2d}
\end{subfigure}
\hspace*{\fill}
\begin{subfigure}{0.32\textwidth}
\includegraphics[width=\linewidth]{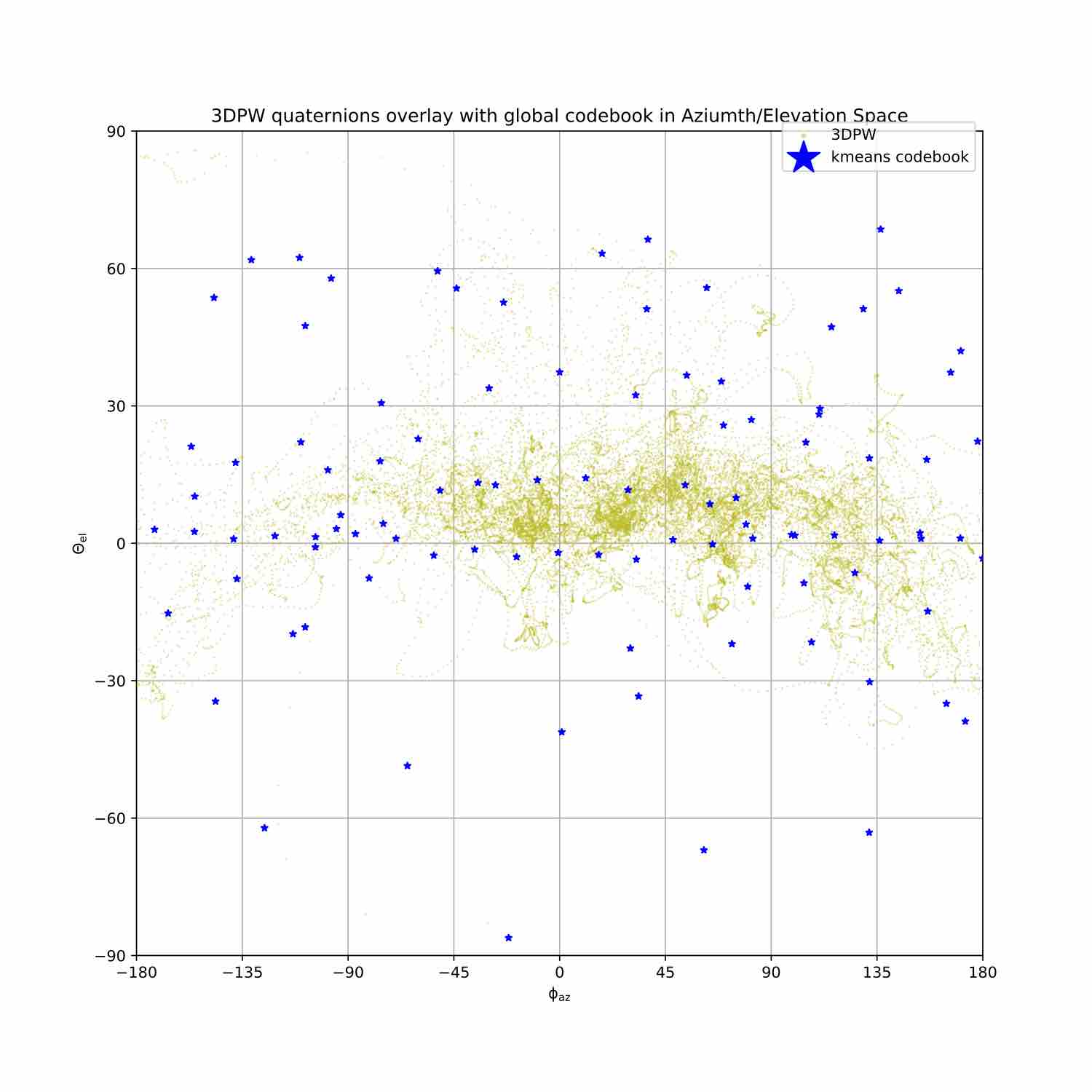}
\caption{\textsc{3DPW}} \label{fig:2e}
\end{subfigure}
\hspace*{\fill}
\begin{subfigure}{0.32\textwidth}
\includegraphics[width=\linewidth]{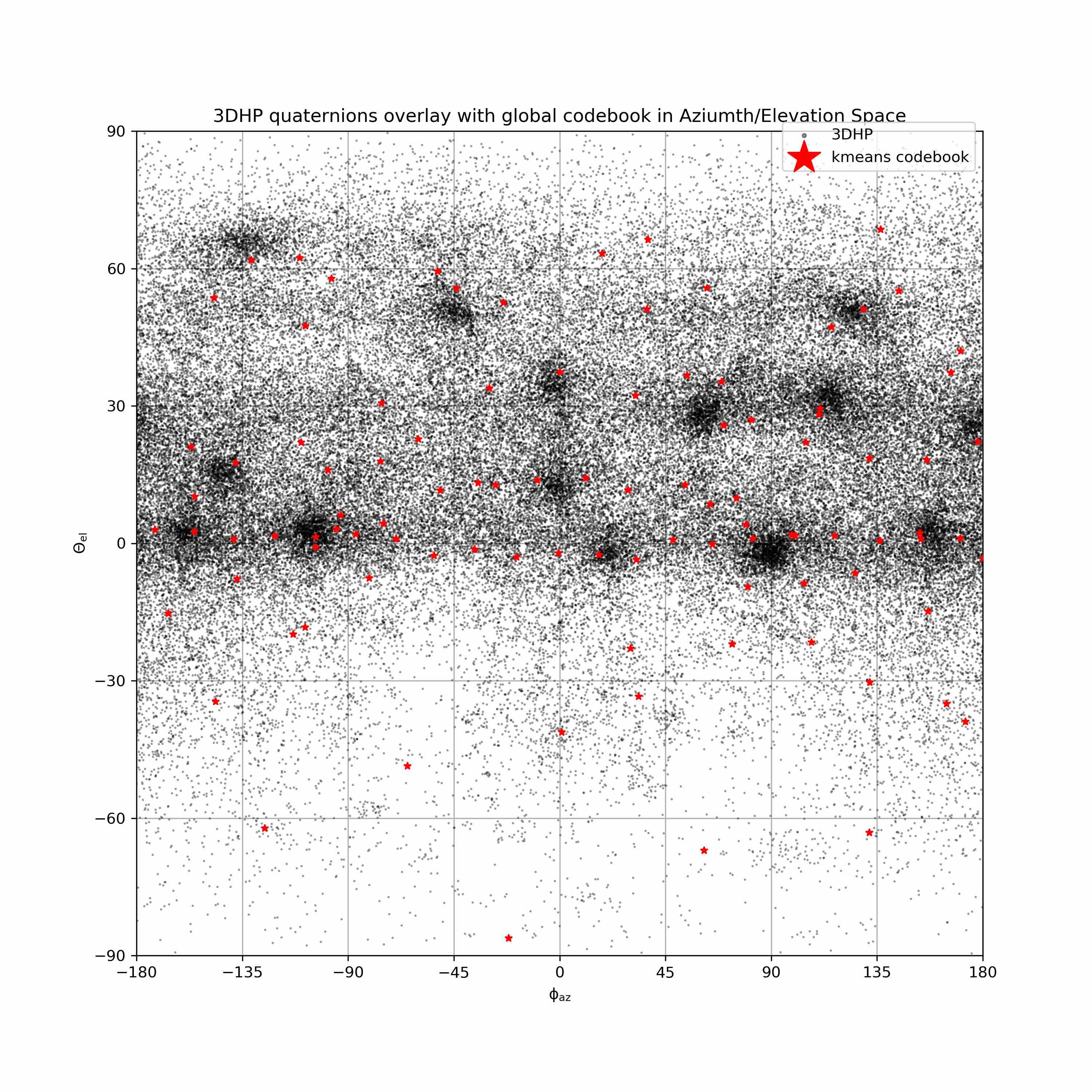}
\caption{\textsc{3DHP}} \label{fig:2f}
\end{subfigure}
\caption{\textbf{a}: Illustration of our body-centered coordinate frame (up
vector, right vector and front vector) relative to a camera-centered coordinate
frame. \textbf{b-f}: Camera viewpoint distribution 
of the 5 datasets overlaid with quaternion cluster centers. Quaternions (rotation 
between body-centered and camera frame) are
sampled from training sets and clustered using k-means.}
\label{fig:body-centeredUFR}
\vspace{-0.15in}
\end{figure*}

\section{Sampled images from five datasets}

\begin{figure*}[t]
\centering
\begin{subfigure}{0.48\textwidth}
\includegraphics[width=\linewidth]{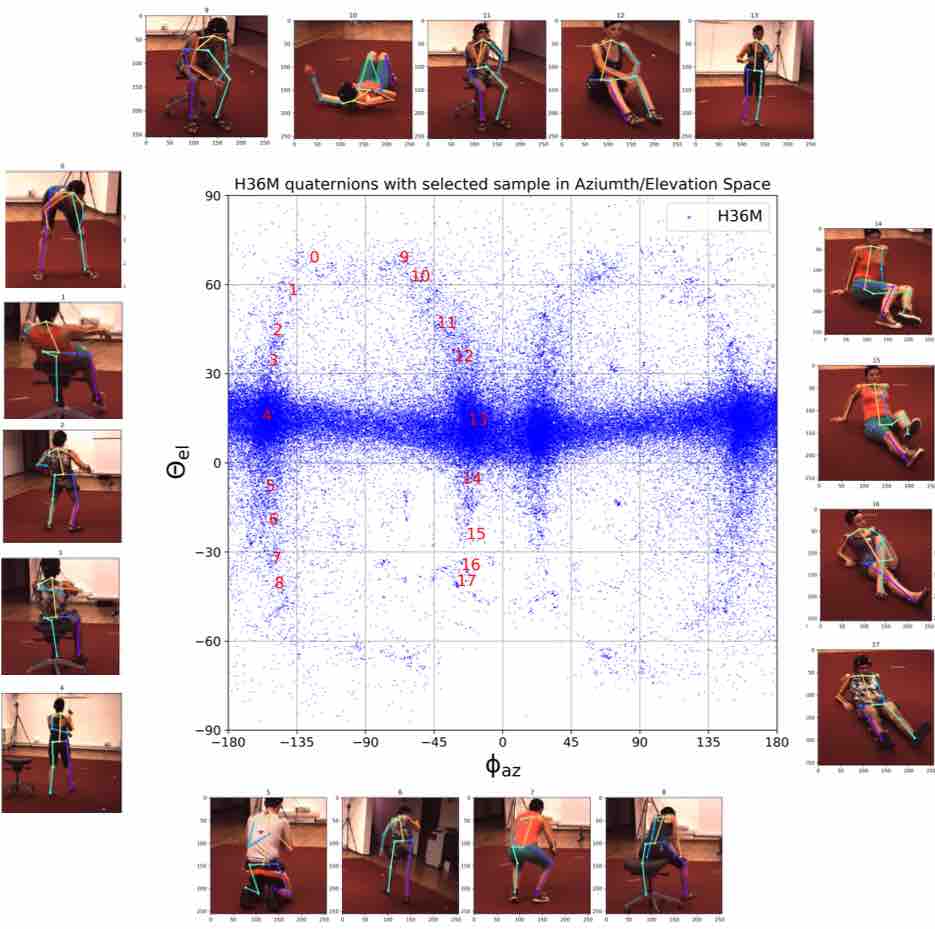}
\caption{H36M with index 0-17} \label{fig:h36m1}
\end{subfigure}
\hspace*{\fill}
\begin{subfigure}{0.48\textwidth}
\includegraphics[width=\linewidth]{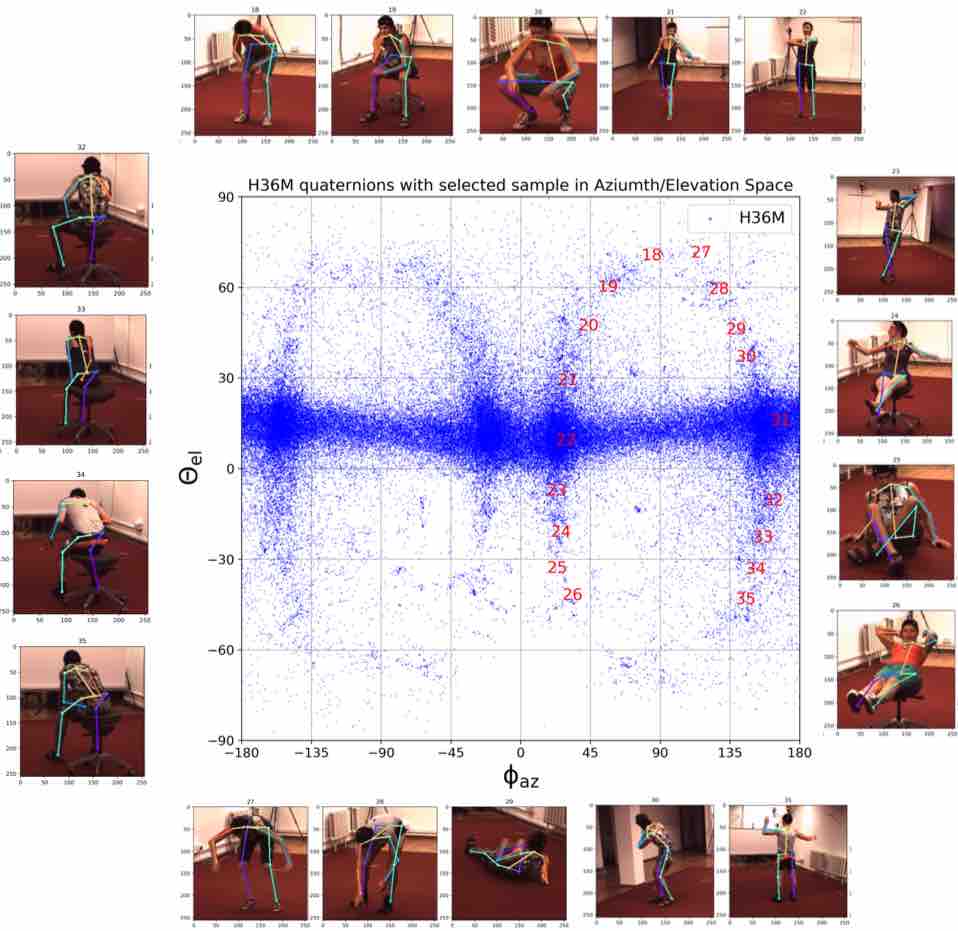}
\caption{H36M with index 18-35} \label{fig:h36m2}
\end{subfigure}
\caption{H36M and sampled images.} 
\label{fig:h36m}
\end{figure*}

\paragraph{Sampled images from H36M} We sample images from the interesting azimuth/elevation pattern from H36M. We can see the images from Fig \ref{fig:h36m1} are facing right while images from Fig \ref{fig:h36m2} are facing left. The index in the azimuth/elevation images corresponds with the index on top of images sampled and placed around the center figure.

\begin{figure*}[t]
\centering
\begin{subfigure}{0.48\textwidth}
\includegraphics[width=\linewidth]{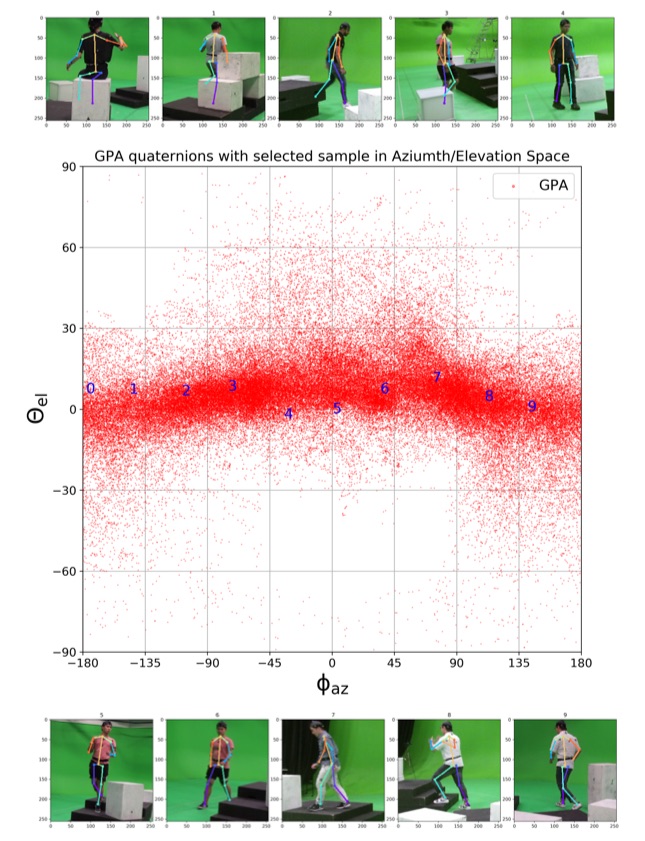}
\caption{GPA with sampled images.} \label{fig:gpa}
\end{subfigure}
\hspace*{\fill}
\begin{subfigure}{0.48\textwidth}
\includegraphics[width=\linewidth]{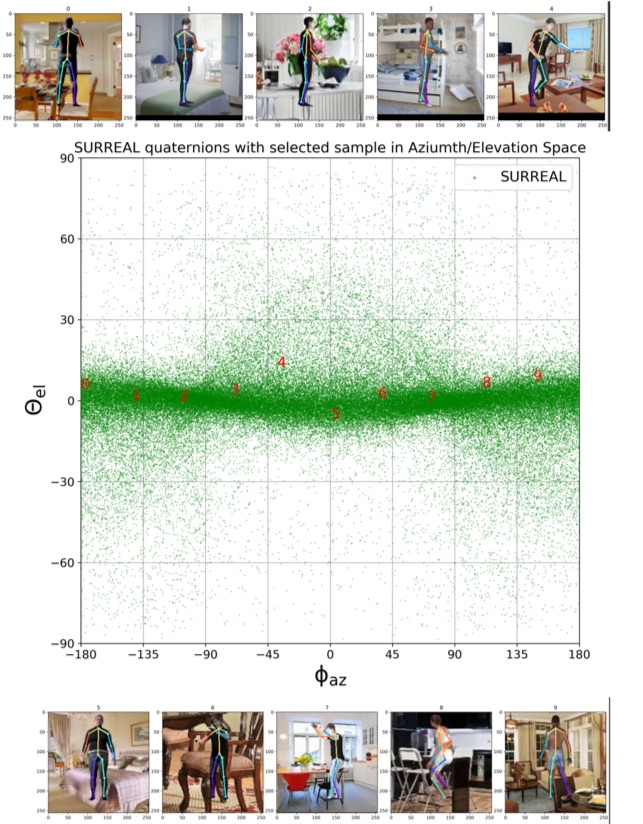}
\caption{SURREAL with sampled images} \label{fig:surreal}
\end{subfigure}
\caption{GPA and SURREAL sampled images.} 
\label{fig:gpasurreal}
\end{figure*}

\paragraph{Sampled images from GPA/SURREAL} We sample images from SURREAL and GPA with uniform azimuth from left to right, and place some randomness on elevation during sampling. We can see the patterns of sampled images from left to right: facing towards back and rotating to facing right, and facing towards the camera, and then facing back again in Fig \ref{fig:gpasurreal}.

\begin{figure*}[t]
\centering
\begin{subfigure}{0.96\textwidth}
\includegraphics[width=\linewidth]{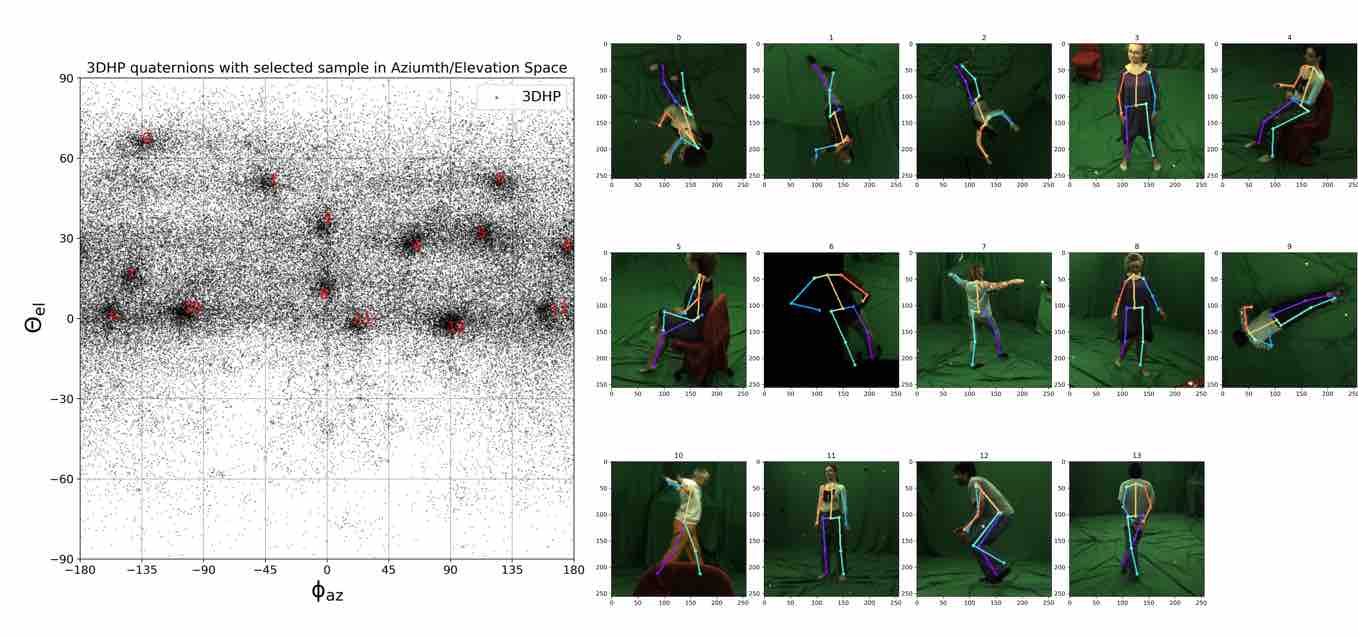}
\caption{3DHP with images sampled from camera center.} \label{fig:d3hpcenter}
\end{subfigure}
\hspace*{\fill}
\begin{subfigure}{0.96\textwidth}
\includegraphics[width=\linewidth]{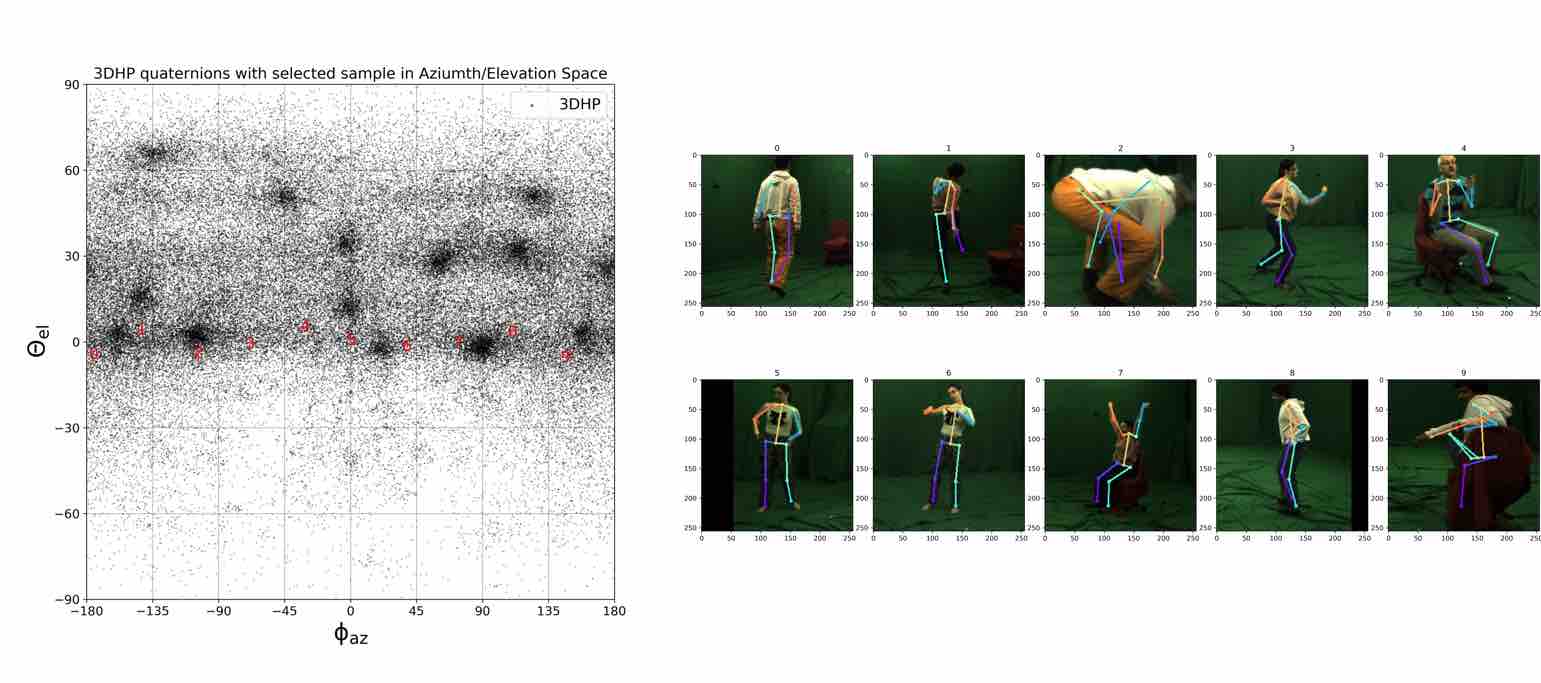}
\caption{3DHP with sampled images in uniform azimuth space.} \label{fig:d3hpazimuth}
\end{subfigure}
\hspace*{\fill}
\begin{subfigure}{0.96\textwidth}
\includegraphics[width=\linewidth]{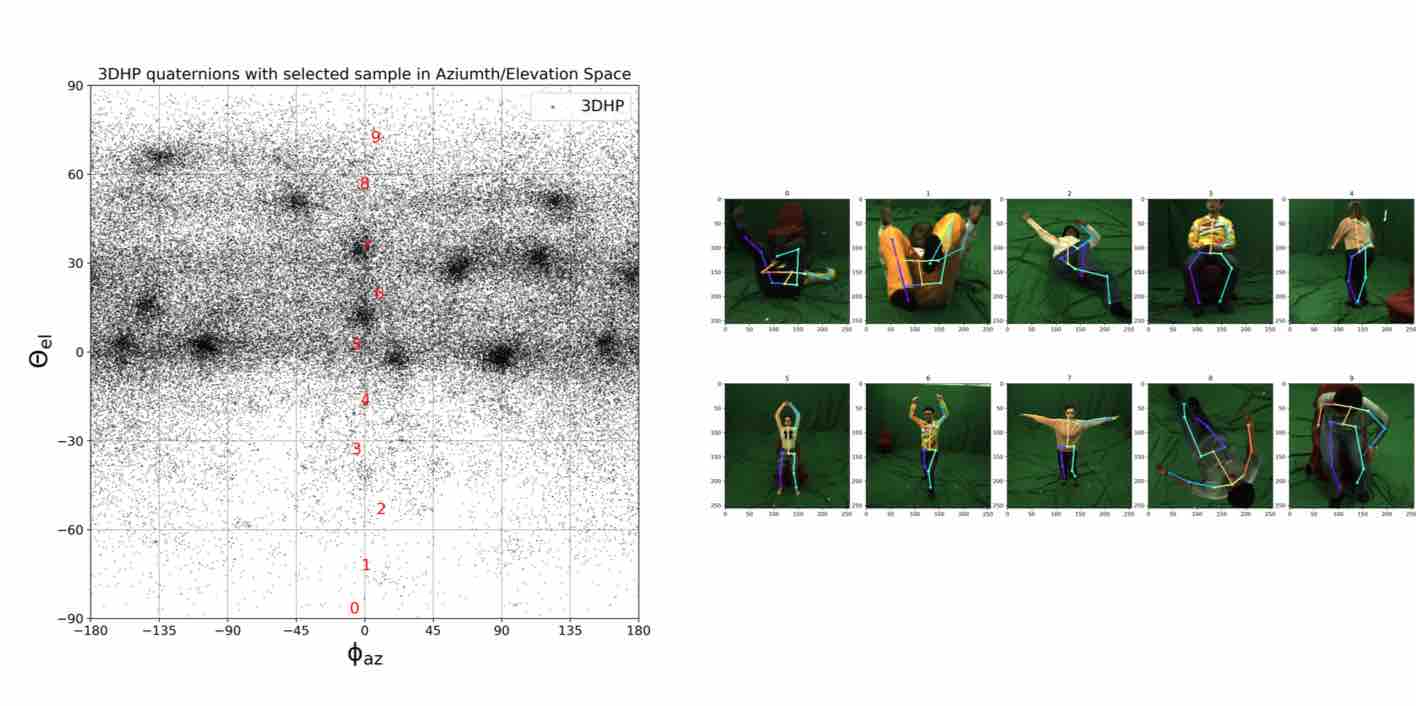}
\caption{3DHP with sampled images in uniform elevation space.} \label{fig:d3hpelevation}
\end{subfigure}
\caption{3DHP sampled images.} 
\label{fig:d3hps}
\end{figure*}

\paragraph{Sampled images from 3DHP} We sample images from 3DHP with uniform azimuth from left to right as shown in Fig~\ref{fig:d3hpazimuth}, uniform elevation from top to down as shown in Fig~\ref{fig:d3hpelevation}, and from camera center as shown in Fig~\ref{fig:d3hpcenter}, during sampling we add some randomness on sampled elevation/azimuth around camera centers.

\begin{figure*}[t]
\centering
\begin{subfigure}{0.96\textwidth}
\includegraphics[width=\linewidth]{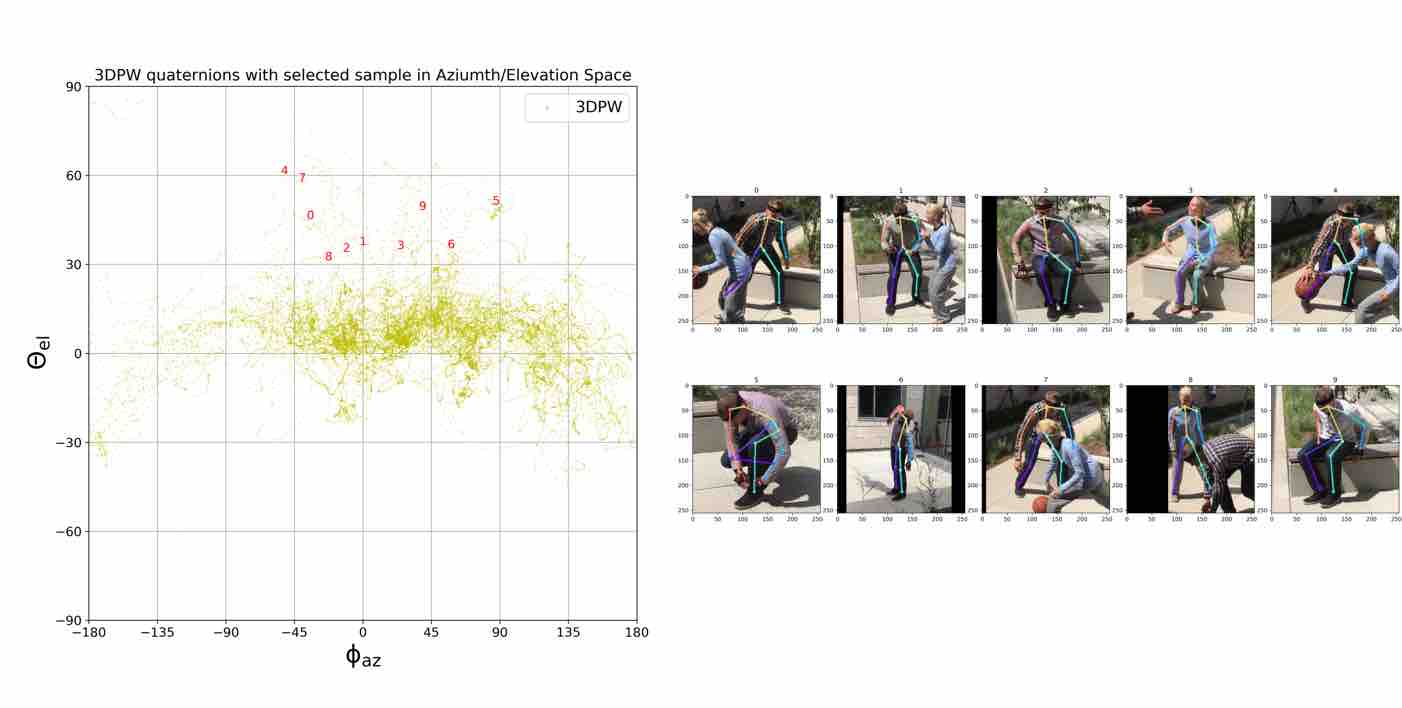}
\caption{3DPW with extreme elevation sampled images.} \label{fig:d3pw1}
\end{subfigure}
\hspace*{\fill}
\begin{subfigure}{0.96\textwidth}
\includegraphics[width=\linewidth]{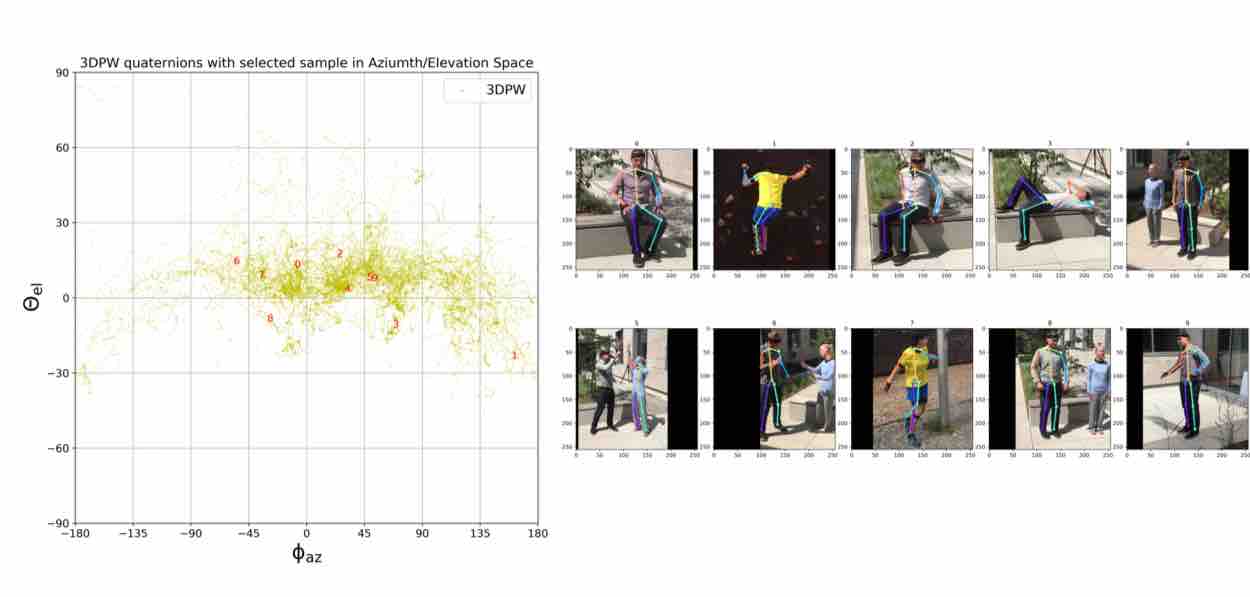}
\caption{3DPW with random sampled images.} \label{fig:d3pw2}
\end{subfigure}
\caption{3DPW sampled images.} 
\label{fig:d3pws}
\end{figure*}

\paragraph{Sampled images from 3DPW} We sample images from 3DPW with extreme elevation as shown in Fig~\ref{fig:d3pw1}, and randomly as shown Fig~\ref{fig:d3pw2}.

\section{Qualitative Results}

\begin{figure*}
\begin{center}
   \includegraphics[width=1\linewidth]{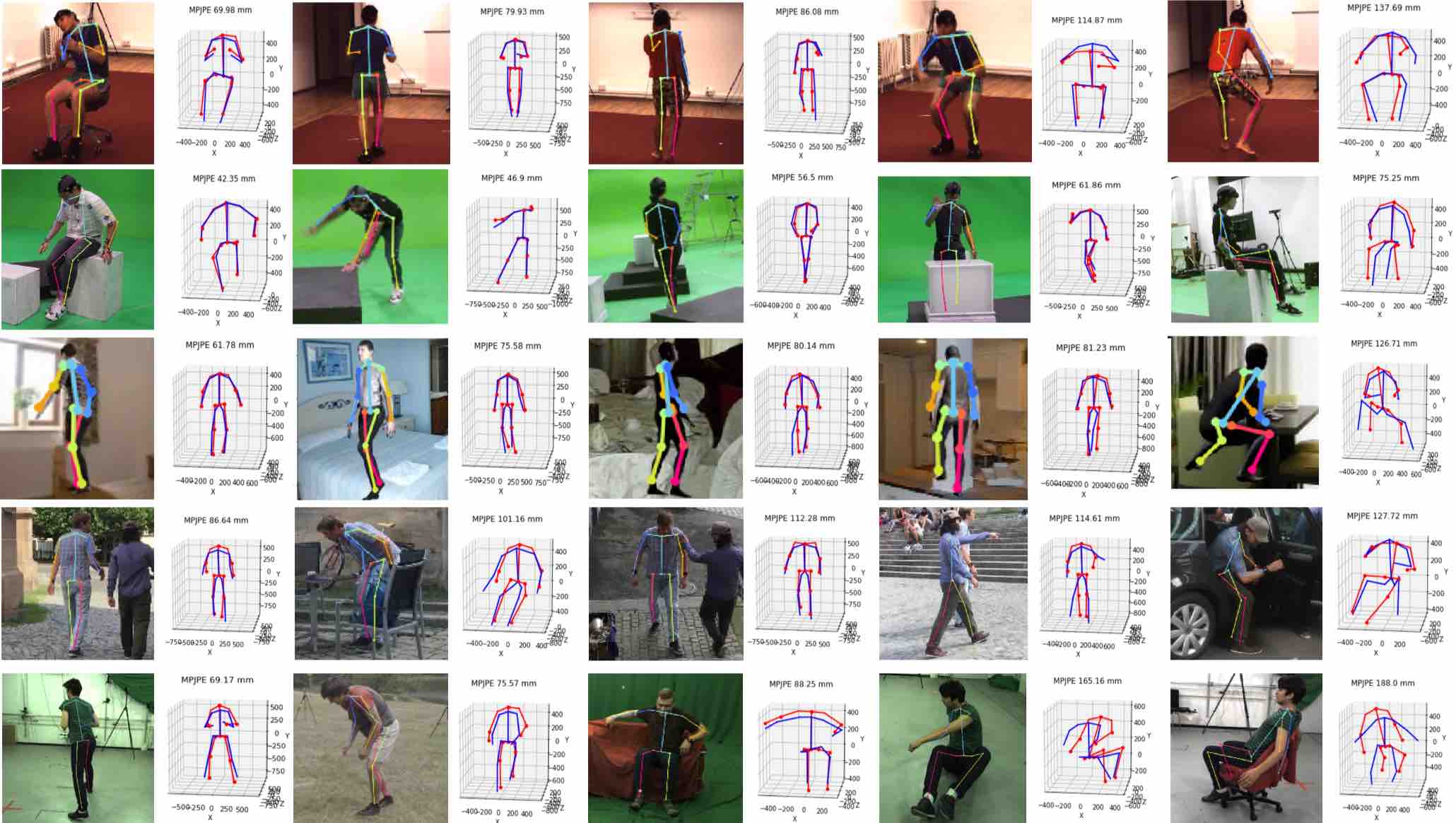}
\end{center}
   \caption{Our prediction on 5 diverse dataset with model trained on GPA dataset. The 2d joints are overlaid with the original image, while the \textcolor{red}{3d prediction (red)} is overlaid with \textcolor{blue}{3d ground truth (blue)}. 3D prediction is \textbf{visualized in body-centered coordinate} rotated by the relative rotation between ground truth root-relative coordinate and body-centered coordinate. From top to bottom are H36M, GPA, SURREAL, 3DPW and 3DHP datasets. We rank the images from left to right in MPJPE increasing order.} 
\label{fig:qualitativeresults1}
\end{figure*}

\begin{figure*}
\begin{center}
   \includegraphics[width=1\linewidth]{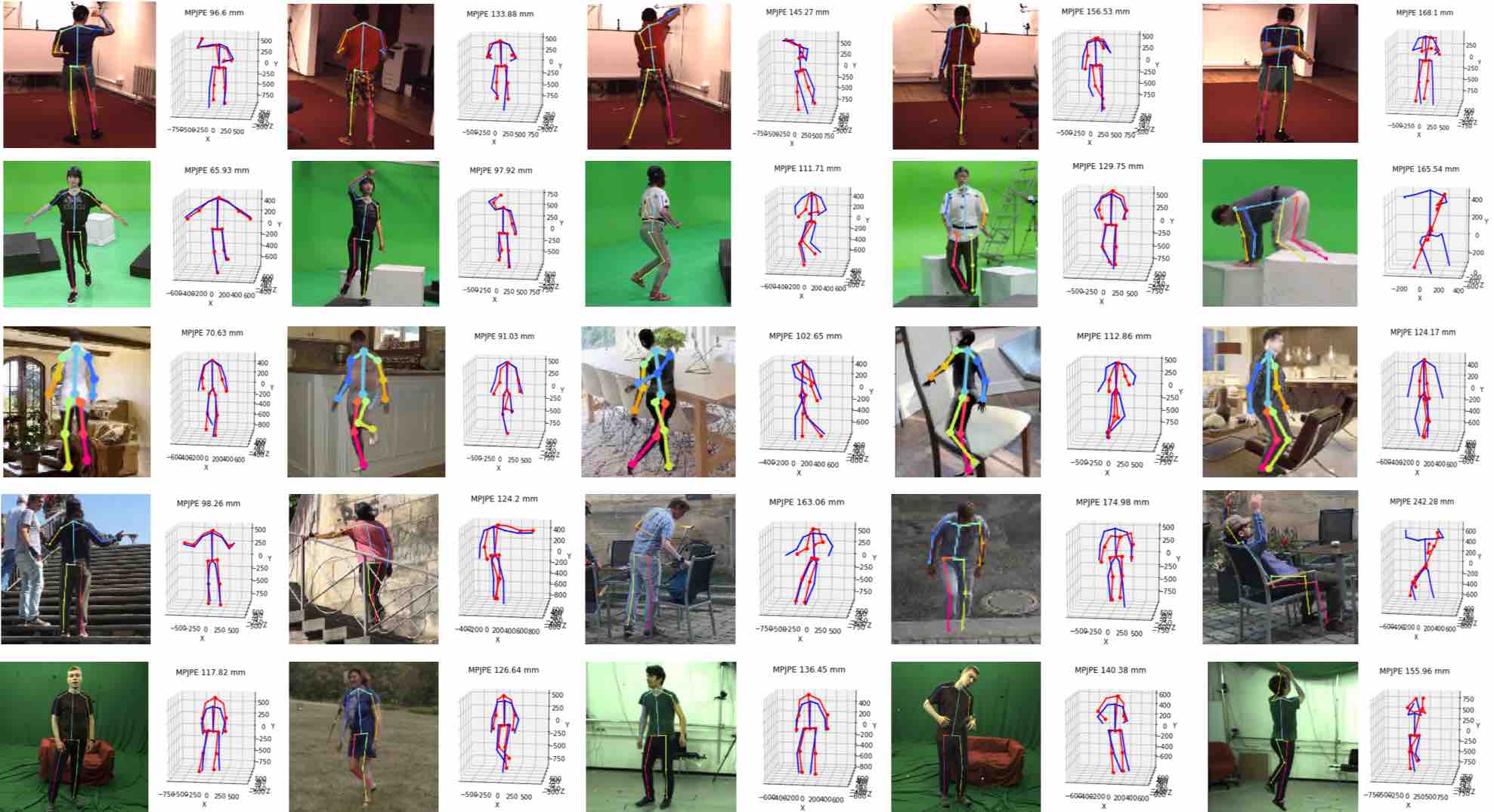}
\end{center}
   \caption{Our prediction on 5 diverse datasets with model trained on SURREAL dataset.} 
\label{fig:qualitativeresults2}
\end{figure*}

\begin{figure*}
\begin{center}
   \includegraphics[width=1\linewidth]{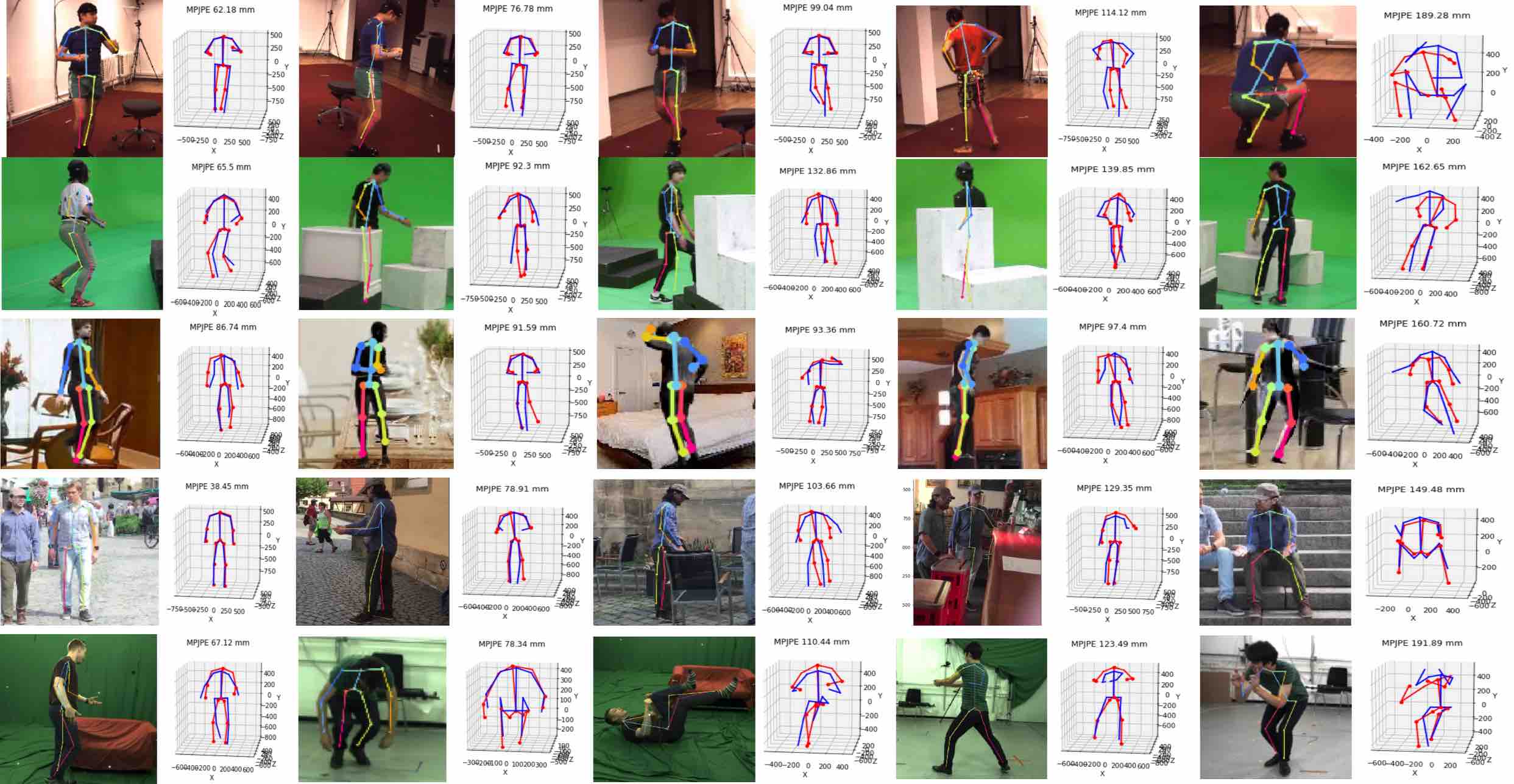}
\end{center}
   \caption{Our prediction on 5 diverse datasets with model trained on 3DPW dataset.} 
\label{fig:qualitativeresults3}
\end{figure*}

\begin{figure*}
\begin{center}
   \includegraphics[width=1\linewidth]{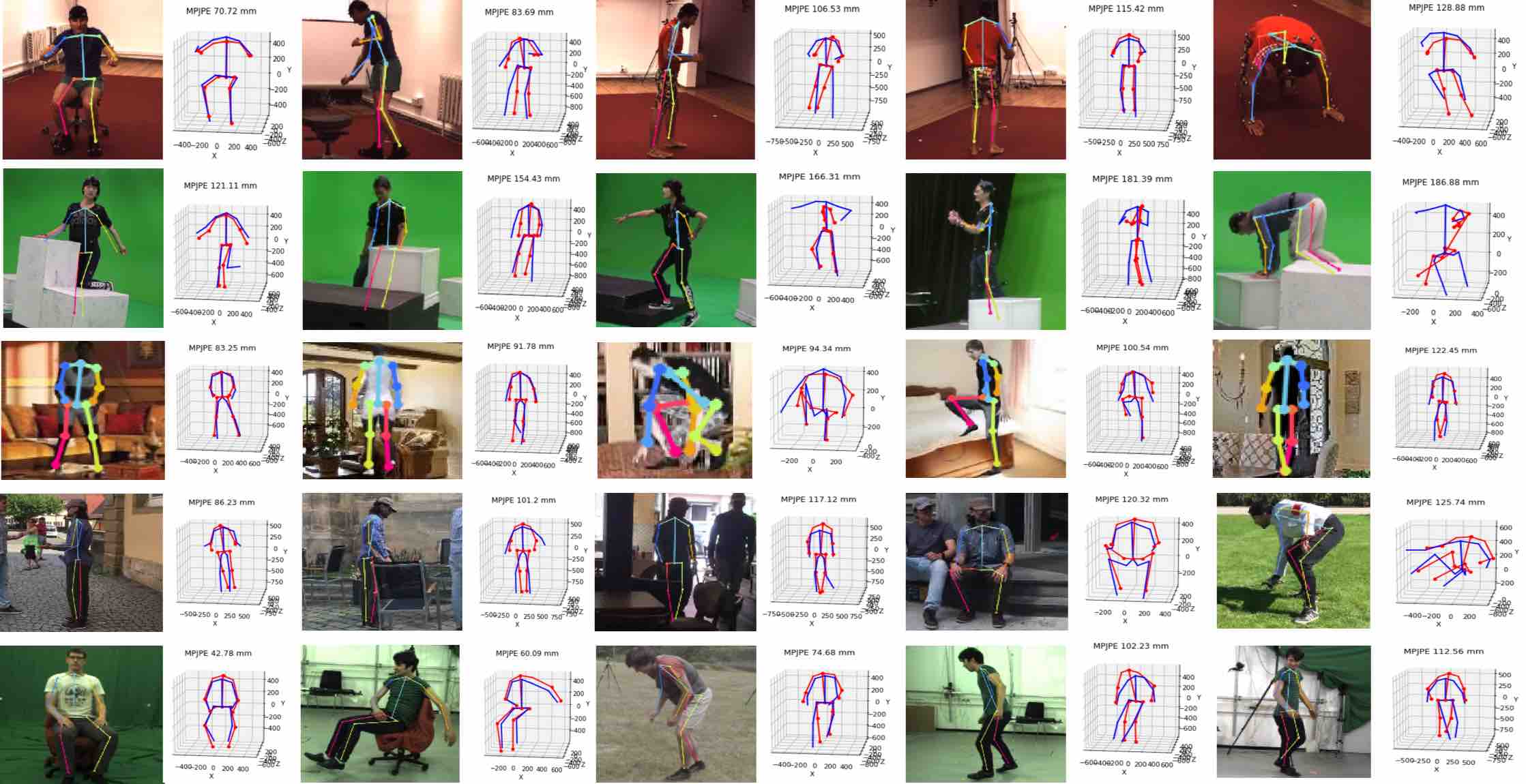}
\end{center}
   \caption{Our prediction on 5 diverse datasets with model trained on  3DHP dataset.} 
\label{fig:qualitativeresults4}
\end{figure*}

\paragraph{Qualitative Results trained on four datasets} We visualize the prediction on the 5 datasets with model trained on \textbf{GPA}, \textbf{SURREAL}, \textbf{3DPW}, \textbf{3DHP} separately on using our proposed method in Fig \ref{fig:qualitativeresults1},\ref{fig:qualitativeresults2},\ref{fig:qualitativeresults3},\ref{fig:qualitativeresults4}. The 2d joint prediction is overlaid with cropped images while the 3d joint prediction is visualized in our proposed body-centered coordinates. From top to bottom are H36M, GPA, SURREAL, 3DPW and 3DHP datasets. We rank the images from left to right in MPJPE increasing order.

\begin{figure*}
\begin{center}
   \includegraphics[width=0.8\linewidth]{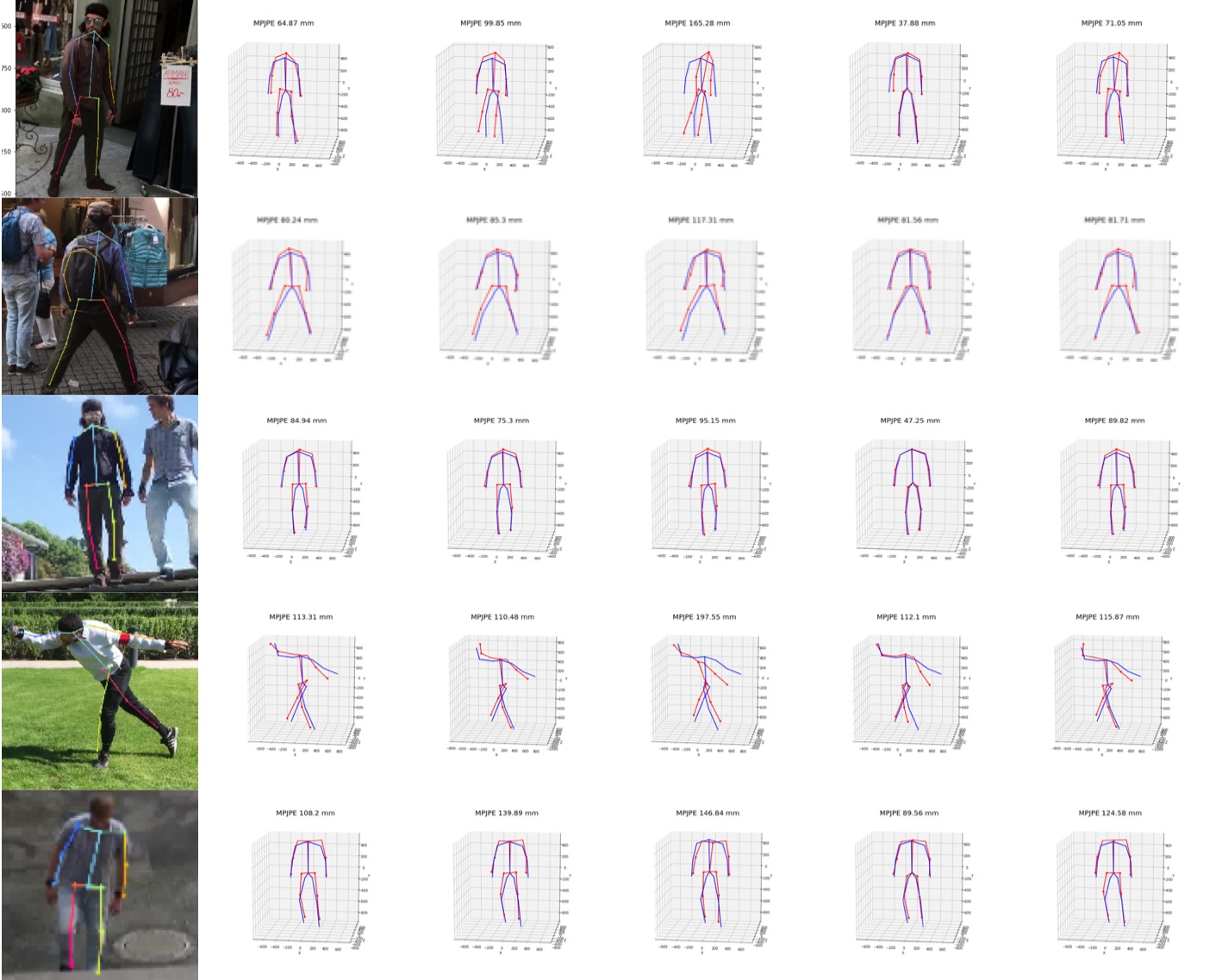}
\end{center}
   \caption{Model trained on 5 models tested on the same images from 3DPW, from left to right (model trained on H36M, GPA, SURREAL, 3DPW, 3DHP).} 
\label{fig:3dpw}
\end{figure*}

\begin{figure*}
\begin{center}
   \includegraphics[width=0.8\linewidth]{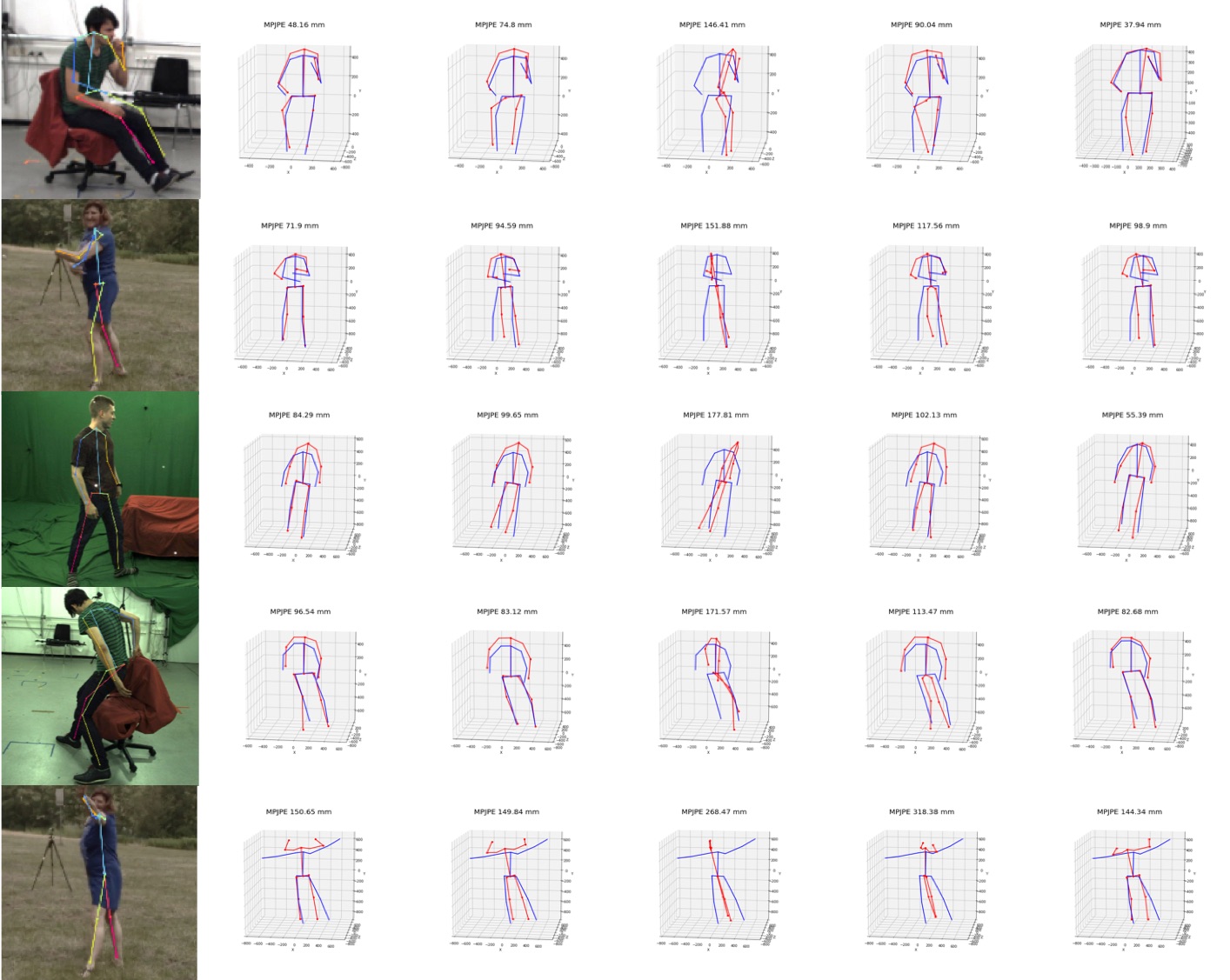}
\end{center}
   \caption{Model trained on 5 models tested on the same images from 3DHP, from left to right (model trained on H36M, GPA, SURREAL, 3DPW, 3DHP).} 
\label{fig:3dhp}
\end{figure*}

\paragraph{Qualitative Results tested on the same images} We further visualize the models trained on 5 datasets, and test on images from the dataset H36M in Fig~\ref{fig:h36m}, GPA in Fig~\ref{fig:gpa},  SURREAL in Fig~\ref{fig:surreal}, 3DPW in Fig~\ref{fig:3dpw} and 3DHP in Fig~\ref{fig:3dhp}.  The results from left to right are models trained on H36M, GPA, SURREAL, 3DPW, and 3DHP.  The RGB images are overlaid with 2d joint prediction from model trained on each dataset.

\begin{figure*}
\begin{center}
   \includegraphics[width=0.8\linewidth]{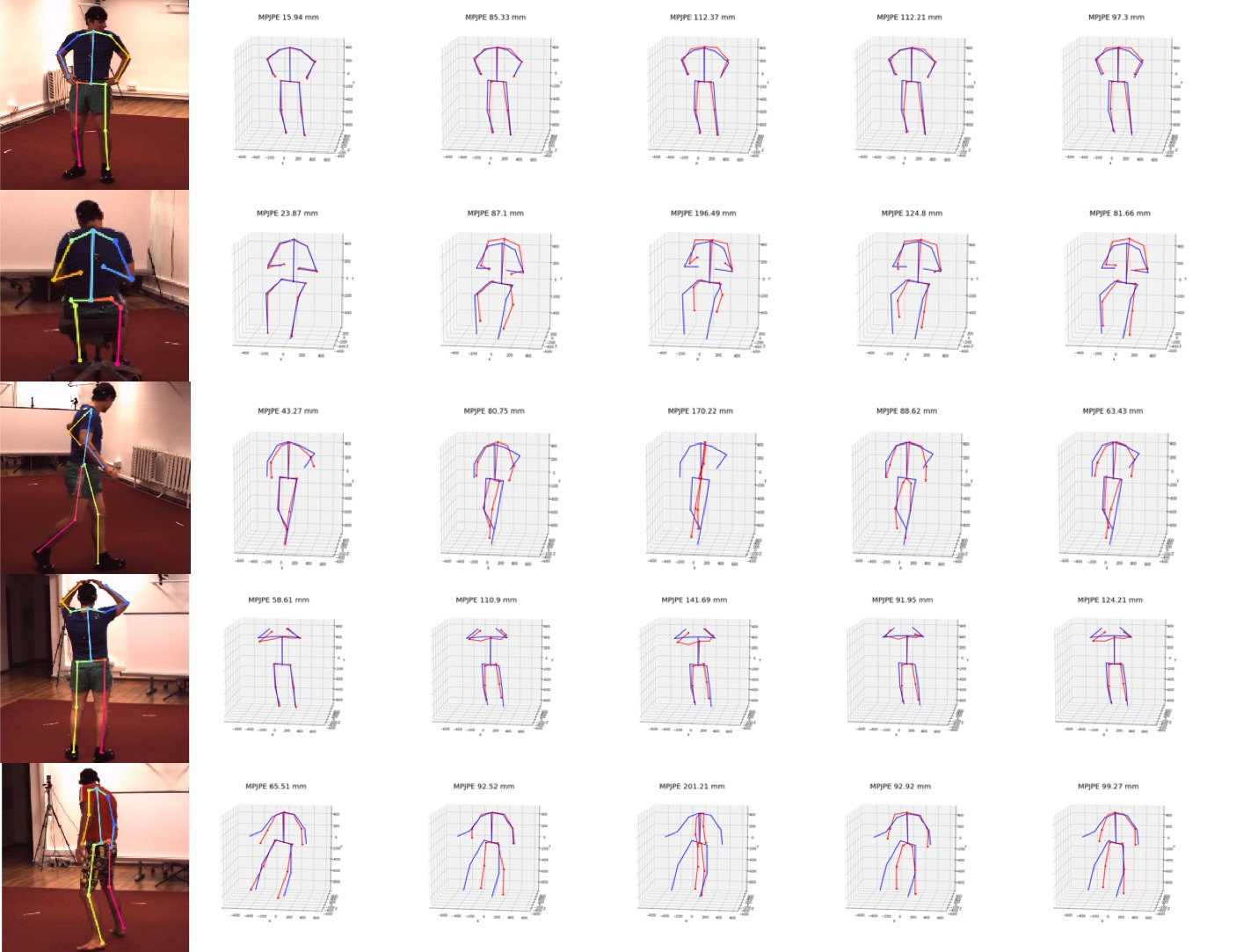}
\end{center}
   \caption{Model trained on 5 models tested on the same images from H36M, from left to right (model trained on H36M, GPA, SURREAL, 3DPW, 3DHP).} 
\label{fig:h36m}
\end{figure*}

\begin{figure*}
\begin{center}
   \includegraphics[width=0.8\linewidth]{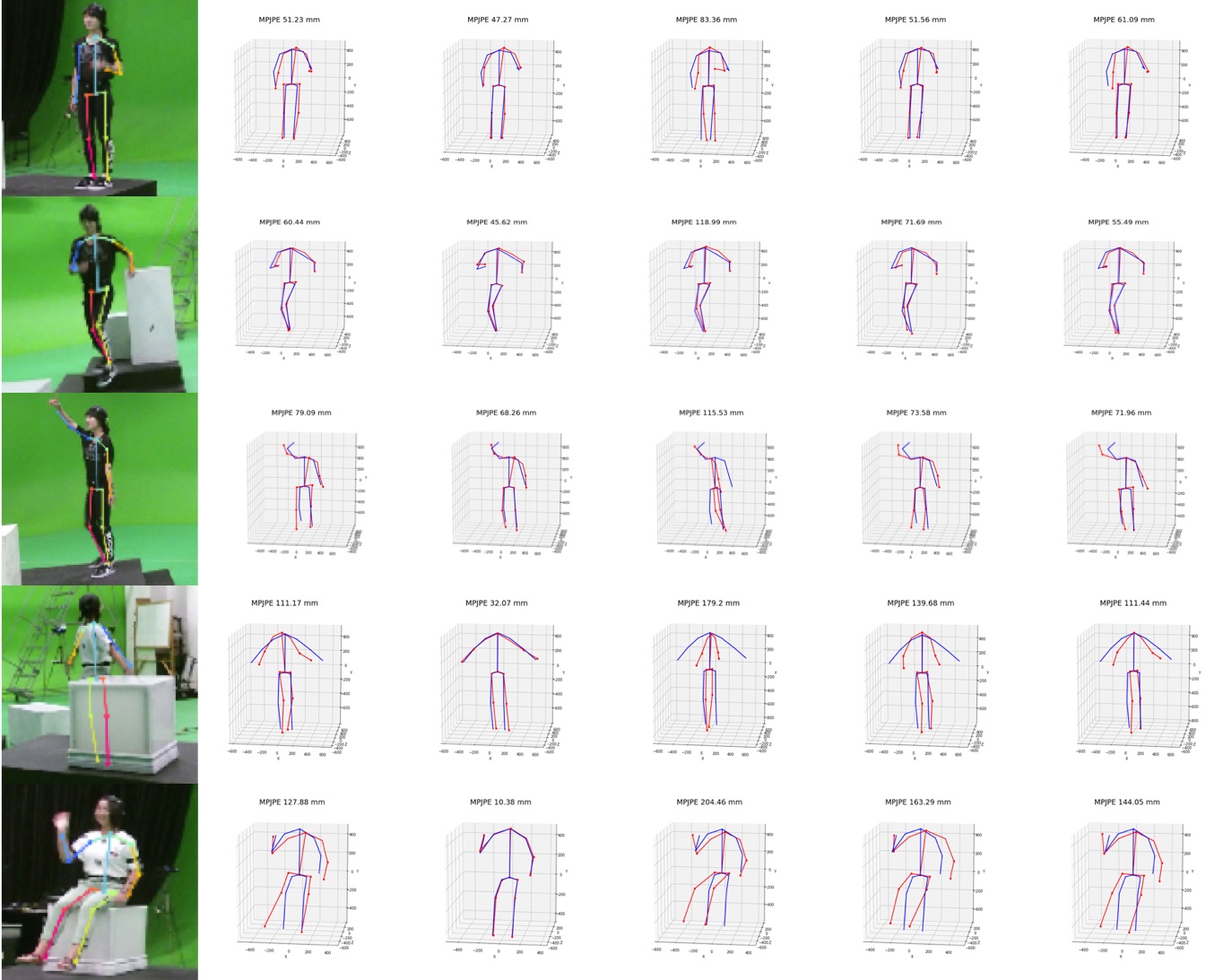}
\end{center}
   \caption{Model trained on 5 models tested on the same images from GPA, from left to right (model trained on H36M, GPA, SURREAL, 3DPW, 3DHP).} 
\label{fig:gpa}
\end{figure*}

\begin{figure*}
\begin{center}
   \includegraphics[width=0.8\linewidth]{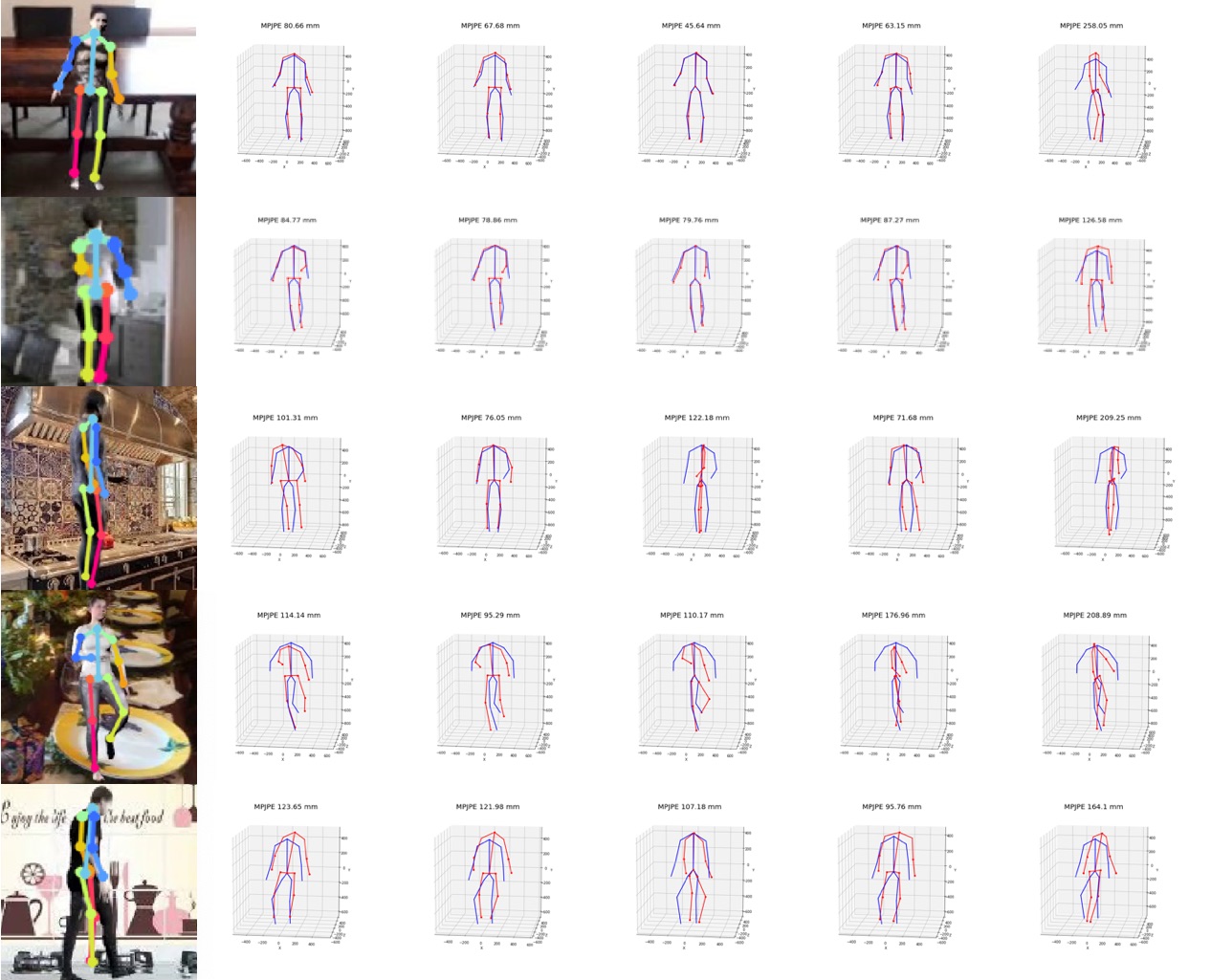}
\end{center}
   \caption{Model trained on 5 models tested on the same images from SURREAL, from left to right (model trained on H36M, GPA, SURREAL, 3DPW, 3DHP).} 
\label{fig:surreal}
\end{figure*}

\end{document}